\documentclass{article}

\usepackage[preprint,nonatbib]{neurips_2020}

\usepackage[utf8]{inputenc} % allow utf-8 input
\usepackage[T1]{fontenc}    % use 8-bit T1 fonts
\usepackage{hyperref}       % hyperlinks
\usepackage{url}            % simple URL typesetting
\usepackage{booktabs}       % professional-quality tables
\usepackage{amsfonts}       % blackboard math symbols
\usepackage{nicefrac}       % compact symbols for 1/2, etc.
\usepackage{microtype}      % microtypography

\usepackage{amsmath}
\usepackage{amssymb}
\usepackage{colortbl}
\usepackage{todonotes}
\usepackage{multicol}
\usepackage[normalem]{ulem} % for strike-out \sout{}

\newcommand{\minisection}[1]{ \noindent {\bf #1}}

\title{RATT: Recurrent Attention to Transient Tasks for Continual Image Captioning}

\author{%
	Riccardo Del Chiaro\thanks{Code for experiments available here: https://github.com/delchiaro/RATT}\\
	MICC, University of Florence\\	
	Florence 50134, FI, Italy\\
	\texttt{riccardo.delchiaro@unifi.it} \\
	
	\And
	Bartłomiej Twardowski \\
	CVC, Universitat Autónoma de Barcelona\\
	08193 Barcelona, Spain \\
	\texttt{bartlomiej.twardowski@cvc.uab.es} \\
	
	\AND
	Andrew D. Bagdanov \\
	MICC, University of Florence\\	
	Florence 50134, FI, Italy\\ 
	\texttt{andrew.bagdanov@unifi.it} \\
	
	\And
	Joost van de Weijer  \\
	CVC, Universitat Autónoma de Barcelona\\
	08193 Barcelona, Spain \\
	\texttt{joost@cvc.uab.es} \\
}

\begin{document}
	
	\maketitle
%	\icmltitle{RATT: Recurrent Attention to Transient Tasks for Continual Image Captioning}
	
	\begin{abstract}
		Research on continual learning has led to a variety of approaches to
		mitigating catastrophic forgetting in feed-forward classification networks.
		Until now surprisingly little attention has been focused on continual learning
		of recurrent models applied to problems like image captioning. In this paper
		we take a systematic look at continual learning of LSTM-based models for image
		captioning. We propose an attention-based approach that explicitly
		accommodates the \emph{transient} nature of vocabularies in continual image
		captioning tasks -- i.e. that task vocabularies are not disjoint. We call our
		method Recurrent Attention to Transient Tasks (RATT), and also show how to
		adapt continual learning approaches based on weight regularization and
		knowledge distillation to recurrent continual learning problems. We apply our
		approaches to incremental image captioning problem on two new continual
		learning benchmarks we define using the MS-COCO and Flickr30 datasets. Our
		results demonstrate that RATT is able to sequentially learn five captioning
		tasks while incurring \emph{no} forgetting of previously learned ones.
	\end{abstract}
	
	\section{Introduction}
	Classical supervised learning systems acquire knowledge by providing them with a
	set of annotated training samples from a task, which for classifiers is a single
	set of classes to learn. This view of supervised learning stands in stark
	contrast with how humans acquire knowledge, which is instead \emph{continual} in
	the sense that mastering new tasks builds upon previous knowledge acquired when
	learning previous ones. This type of learning is referred to as \emph{continual}
	learning (sometimes \emph{incremental} or \emph{lifelong} learning), and
	continual learning systems instead consume a sequence of tasks, each containing
	its own set of classes to be learned. Through a sequence of \emph{learning
		sessions}, in which the learner has access only to labeled examples from the
	current task, the learning system should integrate knowledge from past and
	current tasks in order to accurately master them all in the end. A principal
	shortcoming of state-of-the-art learning systems in the continual learning
	regime is the phenomenon of \emph{catastrophic
		forgetting}~\cite{goodfellow2013empirical,kirkpatrick2017overcoming}: in the
	absence of training samples from previous tasks, the learner is likely to
	\emph{forget} them in the process of acquiring new ones.
	
	Continual learning research has until now concentrated primarily on
	classification problems modeled with deep, feed-forward neural
	networks~\cite{de2019continual,parisi2019continual}. Given the importance of
	recurrent networks for many learning problems, it is surprising that continual
	learning of recurrent networks has received so little
	attention~\cite{Coop2013fixedexpansion,Sodhani2019rnnlll}. A recent study on
	catastrophic forgetting in deep LSTM networks~\cite{Schak2019lstmforgetting}
	observes that forgetting is more pronounced than in feed-forward networks. This
	is caused by the recurrent connections which amplify each small change in the
	weights. In this paper, we consider continual learning for captioning, where a
	recurrent network (LSTM) is used to produce the output sentence describing an
	image. Rather than having access to all captions jointly during training, we
	consider different captioning tasks which are learned in a sequential manner
	(examples of tasks could be captioning of sports, weddings, news, etc).
	
	Most continual learning settings consider tasks that each contain a set
	of classes, and these sets are
	disjoint~\cite{pfulb2019comprehensive,rebuffi2017icarl, serra2018overcoming}. A
	key aspect of continual learning for image captioning is the fact that tasks are
	naturally split into overlapping vocabularies. Task vocabularies might contain
	nouns and some verbs which are specific to a task, however many of the words
	(adjectives, adverbs, and articles) are \emph{shared} among tasks.
	Moreover, the presence of homonyms in different tasks might directly lead to
	forgetting of previously acquired concepts. This \emph{transient} nature of
	words in task vocabularies makes continual learning in image captioning networks
	different from traditional continual learning.
	
	In this paper we take a systematic look at continual learning for image
	captioning problems using recurrent, LSTM networks. We consider three of the
	principal classes of approaches to exemplar-free continual learning:
	weight-regularization approaches, exemplified by Elastic Weight Consolidation
	(EWC)~\cite{kirkpatrick2017overcoming}; knowledge distillation approaches,
	exemplified by Learning without Forgetting (LwF)~\cite{li2017learning}; and
	attention-based approached like Hard Attention to the Task
	(HAT)~\cite{serra2018overcoming}. For each we propose modifications specific to
	their application to recurrent LSTM networks, in general, and more specifically
	to image captioning in the presence of transient task vocabularies.
	
	The contributions of this work are threefold: (1) we propose a new framework and
	splitting methodologies for modeling continual learning of sequential generation
	problems like image captioning; (2) we propose an approach to continual learning
	in recurrent networks based on transient attention masks that reflect the
	transient nature of the vocabularies underlying continual image captioning; and
	(3) we support our conclusions with extensive experimental evaluation on our new
	continual image captioning benchmarks and compare our proposed approach to
	continual learning baselines based on weight regularization and knowledge
	distillation. To the best of our knowledge we are the first to consider
	continual learning of sequential models in the presence of \emph{transient tasks
		vocabularies} whose classes may appear in some learning sessions, then
	disappear, only to reappear in later ones.

	%%%%%%%%%%%%%%%%%%%%%%%%%%%%%%%%%%%%%%%%%%%%%%%%%%%%%%%%%%%%%%%%%%%%%%%%%%%%%%%%%%%%%%%%%%%%%%%
	\section{Related work}
	\label{sec:related}
	
	\minisection{Catastrophic forgetting}. Early works demonstrating the inability
	of networks to retain knowledge from previously task when learning new ones are
	\cite{mccloskey1989catastrophic} and \cite{goodfellow2013empirical}. Approaches
	include methods that mitigate catastrophic forgetting via replay of exemplars
	(iCarl~\cite{rebuffi2017icarl}, EEIL~\cite{castro2018eeil}, and
	GEM~\cite{lopez2017gradient}) or by performing pseudo-replay with GAN-generated
	data~\cite{liu2020generative,shin2017dgr,wu2018memory}. Weight regularization has also been
	investigated~\cite{aljundi2018memory,kirkpatrick2017overcoming,zenke2017continual}.
	Output regularization via knowledge distillation was investigated in
	LwF~\cite{li2017learning}, as well as architectures based on network
	growing~\cite{rusu2016progressive,schwarz2018progress} and attention masking~\cite{mallya2018piggyback,masana2020ternary,serra2018overcoming}. For
	more details we refer to recent surveys on continual
	learning~\cite{parisi2019continual, de2019continual}.

	\minisection{Image captioning}. Modern captioning techniques are inspired 
	by machine translation and usually employ a CNN image encoder and RNN text decoder 
	to ``translate'' images into sentences.
	NIC~\cite{vinyals2015show} uses a pre-trained CNN to encode the image and 
	initialize an LSTM decoder.
	Differently, in \cite{DBLP:journals/corr/MaoXYWY14a} image features are used
	at each time step, while in \cite{donahue2015long} a two-layer LSTM is employed.
	Recurrent latent variable is introduced in \cite{chen2015mind}, encoding
	the visual interpretation of previously-generated words
	and acting as a long-term visual memory during next words generation.
	In~\cite{xu2015show} a spatial attention mechanism is introduced: the model is able to
	focus on specific regions of the image according to the previously generated words.
	\emph{ReviewerNet}~\cite{NIPS2016_6167} also selects in advance which part of the image
	will be attended, so that the decoder is aware of it from the beginning.
	\emph{Areas of Attention}~\cite{pedersoli2017areas} models the dependencies
	between image regions and generated words given the RNN state.
	A visual sentinel is introduced in~\cite{lu2017knowing} to determine, at each
	decoding step, if it is important to attend the visual features.
	The authors of \cite{anderson2018bottom} mixed bottom-up attention (implemented with an object detection
	network in the encoder) and a top-down attention mechanism in the LSTM decoder
	that attend to the visual features of the salient image regions selected by the encoder.
	Recently, transformer-based methods \cite{vaswani2017attention} have been applied to image captioning \cite{huang2019attention, herdade2019image, cornia2020meshed}, which eliminate the LSTM in the decoder.
	
	The focus of this paper is RNN-based captioning architectures and how they are affected by catastrophic forgetting. For more details on image captioning we refer to recent surveys~\cite{Hossain2019capsurvey, li2019visual}.
	
	\minisection{Continual learning of recurrent networks}. A fixed expansion layer
	technique was proposed to mitigate forgetting in RNNs
	in~\cite{Coop2013fixedexpansion}. A dedicated network layer that exploits sparse
	coding of RNN hidden state is used to reduce the overlap of pattern
	representations. In this method the network grows with each new task. A Net2Net
	technique was used for expanding the RNN in~\cite{Sodhani2019rnnlll}. The method
	uses GEM \cite{lopez2017gradient} for training on a new task, but has
	several shortcomings: model weights continue to grow and it must retain
	previous task data in the memory.
	
	Experiments on four synthetic datasets were conducted in~\cite{Schak2019lstmforgetting} to investigate forgetting in LSTM networks. The authors concluded that the LSTM topology has no influence on forgetting. This observations motivated us to take a close look to continual image captioning where the network architecture is more complex and an LSTM is used as a output decoder.

	%%%%%%%%%%%%%%%%%%%%%%%%%%%%%%%%%%%%%%%%%%%%%%%%%%%%%%%%%%%%%%%%%%%%%%%%%%%%%%%%%%%%%%%%%%%%%%%
	\section{Continual LSTMs for transient tasks}
	\label{sec:method}
	
	We first describe our image captioning model and some details of LSTM networks.
	Then we describe how to apply classical continual learning approaches to LSTM
	networks.
	
	\subsection{Image captioning Model}
	\label{sec:captioning}
	We use a captioning model similar to Neural Image Captioning
	(NIC)~\cite{vinyals2015show}. It is an encoder-decoder network that
	``translates'' an image into a natural language description. It is trained
	end-to-end, directly maximizing the probability of correct sequential generation:
	\begin{eqnarray}
	\hat{\theta} &=& \arg\max_\theta \sum_{(I,s)} \log p(s_N|I, s_1, \ldots, s_{N-1}; \theta).
	\end{eqnarray}
	where $s = [s_1, \ldots s_N]$ is the target sentence for image $I$, $\theta$ are
	the model parameters.
	
	The decoder is an LSTM network in which words $s_1, \ldots, s_{n-1}$ are encoded 
	in the hidden state $h_n$ and a linear classifier is used to predict the
	next word at time step $n$:
	
	\begin{tabular}{cc}
		\begin{minipage}{0.51\textwidth}
			\begin{eqnarray}
			x_{0} & = & V\ \mbox{CNN}(I) \label{eq:lstm1}\\ 
			x_n   & = & S\ s_n \label{eq:lstm2}
			\end{eqnarray} 
		\end{minipage} &
		
		\begin{minipage}{0.45\textwidth}
			\begin{eqnarray}
			h_{n} & = & \mbox{LSTM}(x_{n}, h_{n-1}) \label{eq:lstm3}\\
			p_{n+1} & = & C \ h_n \label{eq:lstm4}
			\end{eqnarray}
		\end{minipage}
	\end{tabular}
	
	where $S$ is a word embedding matrix, $s_n$ is the $n$-th word of the ground-truth sentence for image $I$, $C$ is a linear classifier, and $V$ is the visual projection matrix that projects image features from the CNN encoder into the embedding space at time $n=0$.
	
	The LSTM network is defined by the following equations (for which we omit the bias terms): \\
	\begin{tabular}{cc}
		\begin{minipage}{0.51\textwidth}
			\begin{eqnarray}
			\label{eqn:lstm}
			i_n &  = & \sigma ( W_{ix}x_n + W_{ih} h_{n-1}) \\
			o_n &  = & \sigma ( W_{ox}x_n + W_{oh} h_{n-1}) \\
			f_n &  = & \sigma ( W_{fx}x_n + W_{fh} h_{n-1}) \\
			g_n &  = & \tanh ( W_{gx}x_n + W_{gh} h_{n-1})\label{eqn:lstmE}
			\end{eqnarray}
		\end{minipage}
		&
		\begin{minipage}{0.45\textwidth}
			\begin{eqnarray}
			h_n & = & o_n \odot c_n \\
			c_n	& = & f_n \odot c_{n-1} + i_n \odot g_n
			\label{eq:lstm_hidden}
			\end{eqnarray}
		\end{minipage}
	\end{tabular}

	where $\odot$ is the Hadamard (element-wise) product, $\sigma$ the logistic
	function, $c$ the LSTM cell 
	state. The $W$ matrices are the trainable LSTM parameters
	related to input $x$ and hidden state $h$, for each gate
	$i$, $f$, $o$, $g$.
	The loss used to train the network is the sum of the negative log likelihood of the correct word at each step:
	\begin{equation}
	\mathcal{L}(x, s) = - \sum_{n=1}^N \log p_n(s_n).\label{eqn:LSTM_CE}
	\end{equation}
	
	\minisection{Inference.} During training we perform teacher forcing using $n$-th
	word of the target sentence as input to predict word $n+1$. At inference time,
	since we have no target caption, we use the word predicted by the model at the
	previous step $\arg \max \ p_n$ as input to the word embedding matrix $S$.
	
	\subsection{Continual learning of recurrent models}
	\label{sec:continual}
	
	Normally catastrophic forgetting is highlighted in continual learning benchmarks
	by defining tasks that are mutually disjoint in the classes they contain (i.e. no
	class belongs to more than one task). For sequential problems like image
	captioning, however, this is not so easy: sequential learners must classify
	\emph{words} at each decoding step, and a large vocabulary of \emph{common} words
	are needed for any practical captioning task.

	\minisection{Incremental model.} Our models are trained on sequences of
	captioning tasks, each having different vocabularies. For this reason any
	captioning model must be able to enlarge its vocabulary. When a new task arrives
	we add a new column for each new word in the
	classifier and word embedding matrices. The recurrent network remains
	untouched because the embedding projects inputs into the same space. The basic
	approach to adapt to the new task is to fine-tune the network over the new
	training set. To manage the different classes (words) of each task we have two
	possibilities: (1)~Use different classifier and word embedding matrices for each
	task; or (2)~Use a common, growing classifier and a common, growing word
	embedding matrix.
	
	The first option has the advantage that each task can benefit from ad hoc
	weights for the task, potentially initializing from the previous task for the
	common words. However, it also increases decoder network size consistently with each new
	task. The second option has the opposite advantage of keeping the dimension of
	the network bounded, sharing weights for all common words. Because of the nature
	of the captioning problem, many words will be shared and duplicating both word
	embedding matrix and classifier for all the common words seems wasteful. Thus we
	adopt the second alternative. With this approach, the key trick is to
	\emph{deactivate} classifier weights for words not present in the current
	task vocabulary.
	
	We use $\hat{\theta}^t$ to denote optimal weights learned for task $t$ on
	dataset $D_t$. After training on task $t$, we create a new model for task $t+1$
	with expanded weights for classifier and word embedding matrices. We use weights
	from $\hat{\theta}^t$ to initialize the shared weights of the new model.
	
	\subsection{Recurrent continual learning baselines}
	\label{sec:weight-regularization}
	
	We describe how to adapt two common continual learning approaches, one based
	on weight regularization and the other on knowledge distillation. We will use
	these as baselines in our comparison.
	
	\minisection{Weight regularization.} A common method to prevent catastrophic
	forgetting is to apply regularization to important model weights before
	proceeding to learn a new
	task~\cite{aljundi2018memory,chaudhry2018riemannian,kirkpatrick2017overcoming,zenke2017continual}.
	Such methods can be directly applied to recurrent models with little effort. We
	choose Elastic Weight Consolidation (EWC)~\cite{serra2018overcoming} as a
	regularization-based baseline. The key idea of EWC is to limit change to model
	parameters vital to previously-learned tasks by applying a quadratic penalty to
	them depending on their importance. Parameter importance is estimated using a
	diagonal approximation of the Fisher Information Matrix. The additional loss function we
	minimize when learning task $t$ is:
	\begin{equation}
	\label{eqn:ewc_loss}
	\mathcal{L}^t_{\mathrm{EWC}}(x, S; \theta^t) =  \mathcal{L}(x, S) + \lambda 
	\textstyle    % added to gain a line
	\sum_i\frac{1}{2} \  F^{t-1}_i(\theta^t_i - \hat{\theta}^{t-1}_{i})^2,
	\end{equation}
	where $\hat{\theta}^{t-1}$ are the estimated model parameters for the previous
	task, $\theta^{t}$ are the model parameters at the current task $t$,
	$\mathcal{L}(x, S)$ is the standard loss used for fine-tuning the network on
	task $t$, $i$ indexes the model parameters shared between tasks $t$ and $t-1$,
	$F^{t-1}_i$ is the $i$-th element of a diagonal approximation of the Fisher
	Information Matrix for model after training on task $t-1$, and $\lambda$ weights
	the importance of the previous task. We apply
	Eq.~\ref{eqn:ewc_loss} to all trainable weights. Due to the transient nature of
	words across tasks, we do not expect weight regularization to be optimal since
	some words are shared and regularization limits the plasticity needed to
	adjust to a new task.

	\minisection{Recurrent Learning without Forgetting}.
	\label{sec:LwF}
	We also apply a knowledge distillation~\cite{hinton2015distilling} approach
	inspired by Learning without Forgetting (LwF)~\cite{li2017learning} on the LSTM
	decoder network to prevent catastrophic forgetting.
	The model after training task $t-1$ is used as a teacher network when
	fine-tuning on task $t$. The aim is to let the new network freely learn how to
	classify new words appearing in task $t$ while keeping stable the predicted
	probabilities for words from previous tasks.
	
	To do this, at each step $n$ of the decoder network the previous decoder is also
	fed with the data coming from the new task $t$. Note that the input to the LSTM
	at each step $n$ is the embedding of the $n-$th word in the target caption, and
	the same embedding is given as input to both teacher and student networks -- i.e.
	the student network's embedding of word $n$ is also used as input for the
	teacher, while each network uses its own hidden state $h_{n-1}$ and cell state
	$c_{n-1}$ to decode the next word. At each decoding step, the output
	probabilities $p^{t, <t}_{n+1}$ from the student network LSTM corresponding to
	words present in tasks $1, \dots, t-1$ are compared with the those predicted by
	the teacher network, $p^{t-1, <t}_{n+1}$. A distillation loss ensures that the
	student network does not deviate from the teacher:
	\begin{eqnarray}
	\mathcal{L}^{t}_{\mathrm{LwF}}(\hat{p}^t, \hat{p}^{t-1}) & = & - 
	\textstyle 
	\sum_n H(\gamma(p^{t, <t}_n), \gamma(p^{t-1, <t}_n))
	\end{eqnarray}
	where $\gamma(\cdot)$ rescales a probability vector $p$ -with temperature parameter $T$. This loss 
	is combined with the LSTM 
	training loss (see Eq.~\ref{eqn:LSTM_CE}).
	Note that differently from~\cite{li2017learning}, we do not fine-tune the classifier of the old network because we use a single, incremental word classifier.

	%%%%%%%%%%%%%%%%%%%%%%%%%%%%%%%%%%%%%%%%%%%%%%%%%%%%%%%%%%%%%%%%%%%%%%%%%%%%%%%%%%%%%%%%%%%%%%%%%%%%%%%%%%%%%%%%%

	\section{Attention for continual learning of transient tasks}
\begin{figure}
		\resizebox{1\textwidth}{!}{
			\includegraphics{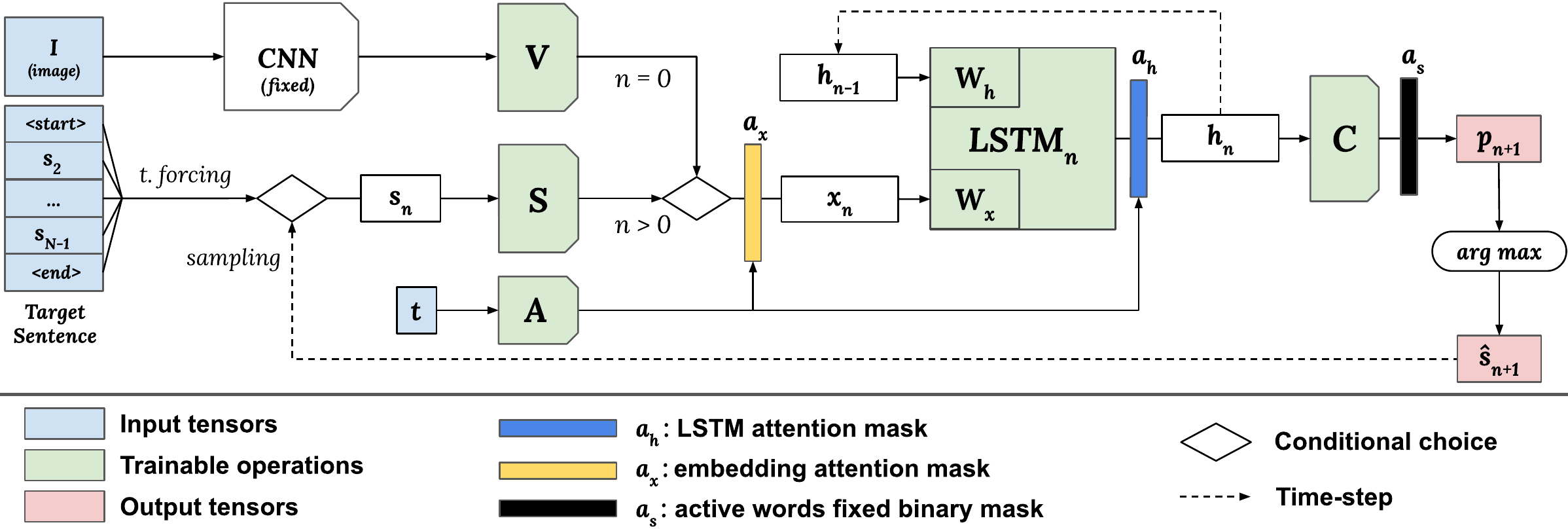}}
		\vspace{-0.5cm}
		\caption{
			\label{fig:approaches}
			Recurrent Attention to Transient Tasks (RATT). See section~\ref{sec:RAT} for a detailed description of each component of the continual captioning network.
		}
	\end{figure}

	Inspired by the Hard Attention to the Task (HAT)
	method~\cite{serra2018overcoming}, we developed an attention-based technique
	applicable to recurrent networks. We name it \textit{Recurrent Attention to
		Transient Tasks (RATT)}, since it is specifically designed for recurrent networks with
	task transience. The key idea is to use an attention mechanism to allocate a
	portion of the activations of each layer to a \emph{specific} task
	$t$. An overview of RATT is provided in figure~\ref{fig:approaches}.

	\minisection{Attention masks}. The number of neurons used for a task is limited by two task-conditioned attention masks: embedding attention $a_x^t\in [0, 1]$ and hidden state attention $a_h^t\in [0, 1]$. These are computed with a sigmoid activation $\sigma$ and a positive scaling factor $s$ according to:
	\vspace{-0.35cm}
	
	\begin{eqnarray}
	a_x^t= \sigma(s A_x t^T) \;,\; a_h^t= \sigma(s A_h t^T),\label{eqs:embedding}
	\end{eqnarray}
	where $t$ is a one-hot task vector, and $A_x$ and $A_h$ are embedding matrices.  Next to the two attention mask, we have a vocabulary mask $a_s^t$ which is a binary mask identifying the words of the vocabulary used in task $t$: $a^t_{s,i}=1$ if word $i$ is part of the vocabulary of task $t$ and is zero otherwise. The forward pass (see Eqs.\ref{eq:lstm1} and \ref{eq:lstm4}) of the network is modulated with the attention masks according to: 
	\vspace{-0.35cm}
	
	\begin{tabular}{cc}
		\begin{minipage}{0.51\textwidth}
			\begin{eqnarray}
			\bar{x}_0&=&x_0 \odot a_x^t\\
			\bar{x}_n&=&x_n \odot a_x^t
			\end{eqnarray} 
		\end{minipage} &
		\begin{minipage}{0.45\textwidth}
			\begin{eqnarray}
			\bar{h}_n&=&h_n \odot a_h^t\\
			\bar{p}_{n+1}&=& p_{n+1} \odot a_s^t
			\end{eqnarray}
		\end{minipage}
	\end{tabular}
	
	Attention masks act as an inhibitor
	when their value is near $0$. The main idea is to learn attention masks
	during training, and as such learn a limited set of neurons for each task.
	Neurons used in previous tasks can still be used in subsequent ones,
	however the weights which were important for previous tasks have reduced
	plasticity (depending on the amount of attention to for previous
	tasks).

	\label{sec:RAT}
	
	\minisection{Training}. For training we define the cumulative forward mask
	as:
	\begin{eqnarray}
	a_x^{<t} &=& \max(a_x^{t-1}, a_x^{<t-1}),
	\end{eqnarray}
	$a_h^{<t}$ and $a_s^{<t}$ are similarly defined. We now define the following backward masks which have the dimensionality of the weight matrices of the network and are used to selectively backpropagate the gradient to the LSTM layers:
	\begin{eqnarray}
	B^t_{h, ij} = 1 - \min(a_{h,i}^{<t}, a_{h,j}^{<t})\;,\;B^t_{x, ij}= 1 - \min(a_{h,i}^{<t}, a_{x,j}^{<t})
	\end{eqnarray}
	Note that we use $a_{h,i}$ refer to the i-th element of vector $a_h$, etc. The backpropagation with learning rate $\lambda$ is then done according to
	\begin{eqnarray}
	W_{h}\leftarrow W_{h} - \lambda B^t_{h} \odot \frac{\partial \mathcal{L}^t}{\partial W_{h}}\;,\;
	W_{x}\leftarrow W_{x} - \lambda B^t_{x} \odot \frac{\partial \mathcal{L}^t}{\partial W_{x}}.
	\end{eqnarray}
	The only difference from standard backpropagation are the backward matrices $B$ which prevents the gradient from changing those weights that were highly attended in previous tasks. The backpropagation updates to the other matrices in Eqs.~\ref{eqn:lstm}-\ref{eqn:lstmE} are similar (see Suppl. Materials).
	
	Other than~\cite{serra2018overcoming} we also define backward masks for the word embedding matrix $S$, the linear classifier $C$, and the image-projection matrix $V$:
	\begin{eqnarray}
	B^t_{S, ij}= 1 - \min(a_{x,i}^{<t},a_{s,j}^{<t})\;,\;B^t_{C, ij}= 1 - \min(a_{s,i}^{<t},a_{h,j}^{<t}) \;,\; B^t_{V, ij} = 1 - a_{x,i}^{<t},
	\end{eqnarray}
	and the corresponding backpropagation updates: 
	\begin{eqnarray}
	S\leftarrow S - \lambda B^t_{S} \odot \frac{\partial \mathcal{L}^t}{\partial S}\;,\;
	C\leftarrow C - \lambda B^t_{C} \odot \frac{\partial \mathcal{L}^t}{\partial C}\;,\;
	V\leftarrow V - \lambda B^t_{V} \odot \frac{\partial \mathcal{L}^t}{\partial V}.
	\end{eqnarray}
	The backward mask $B^t_{V}$ modulates the backpropagation to the image features. Since we do not define a mask on the output of the fixed image encoder, this is only defined by $a_{x}^{<t}$.

	Linearly annealing the scaling parameter $s$, used in Eq.~\ref{eqs:embedding}, during training (like~\cite{serra2018overcoming}) was found to be beneficial. We apply $s = \frac{1}{s_{max}} + \left(s_{max} - \frac{1}{s_{max}} \right) \frac{b-1}{B-1}$
	where $b$ is the batch index and $B$ is the total number of batches for the epoch. We used $s_{max} = 2000$ and $s_{max} = 400$ for experiments on Flickr30k and MS-COCO, respectively.
	
	The loss used to promote low network usage and to keep some neurons available for future tasks is:
	\begin{equation}
	\mathcal{L}^t_{a} = \frac
	{ \sum_{i} a^t_{x, i} (1-a^{<t}_{x, i})}
	{ \sum_{i} (1-a^{<t}_{x, i})}+\frac
	{ \sum_{i} a^t_{h, i} (1-a^{<t}_{h, i})}
	{ \sum_{i} (1-a^{<t}_{h, i})}.
	\end{equation}
	
	This loss is combined with Eq.~\ref{eqn:LSTM_CE} for training. The loss
	encourages attention to only a few new neurons. However, tasks can attend to
	previously attended neurons without any penalty. This encourages forward
	transfer during training. If the attention masks are binary, the system would not
	suffer from any forgetting, however it would lose its backward transfer ability.
	
	Differently than~\cite{serra2018overcoming}, when computing $B_S^t$ we take into account the recurrency of the network, considering the classifier $C$ to be the previous layer of $S$. In addition, our output masks $a_s$ allow for overlap to model the transient nature of the output vocabularies, whereas~\cite{serra2018overcoming} only considers non-overlapping classes for the various tasks.

	%%%%%%%%%%%%%%%%%%%%%%%%%%%%%%%%%%%%%%%%%%%%%%%%%%%%%%%%%%%%%%%%%%%%%%%%%%%%%%%%%%%%%%%%%%%%%%%
	\section{Experimental results}
	\label{sec:experiments}
	All experiments use the same architecture: for the encoder network we used
	ResNet151~\cite{he2016deep} pre-trained on
	ImageNet~\cite{russakovsky2015imagenet}. Note that the image encoder is frozen and is not trained during continual learning, as is common in many image captioning systems.
	The decoder consists of the word
	embedding matrix $S$ that projects the input words into a 256-dimensional
	space, an LSTM cell with hidden size 512 that takes the word (or image
	feature for the first step) embeddings as input, and a final fully connected layer $C$ that take as input the hidden state $h_n$ at each LSTM step $n$ and outputs a probability distribution $p_{n+1}$ over the $|V^t|$ words in the vocabulary for current task $t$.
	
	We applied all techniques on the Flickr30K~\cite{flickrentitiesijcv} and
	MS-COCO~\cite{lin2014mscoco} captioning datasets (see next section for task
	splits). All experiments were conducted using PyTorch, networks were trained
	using the Adam~\cite{kingma2014adam} optimizer, all hyperparameters were tuned
	over validation sets.
	Batch size, learning rate and max-decode length for evaluation
	were set, respectively, to 128, 4e-4, and 26 for MS-COCO, and 32, 1e-4 and 40
	for Flickr30k. These differences are due to the size of the training set and
	by the average caption lengths in the two datasets.
	
	Inference at test time is \emph{task-aware} for all methods. For EWC and LwF
	this means that we consider only the word classifier outputs corresponding to
	the correct task, and for RATT that we use the fixed output masks for the correct task. 
	All metrics where computed using the nlg-eval toolkit~\cite{sharma2017nlgeval}. 
	Models where trained for a
	fixed number of epochs and the best model according to BLEU-4 performance on the
	validation set were chosen for each task. When proceeding to the next task, the
	best model from the previous task were used as a starting point.

	\subsection{Datasets and task splits}
	For our experiments we use two different captioning datasets:
	MS-COCO~\cite{flickrentitiesijcv} and Flickr30k~\cite{lin2014mscoco}. We split
	MS-COCO into tasks using a \emph{disjoint visual categories} procedure. For
	this we defined five tasks based on disjoint MS-COCO super-categories containing
	related classes (\emph{transport}, \emph{animals}, \emph{sports}, \emph{food} and \emph{interior}).
	For Flickr30K we instead used an \emph{incremental visual categories} procedure.
	Using the visual entities, phrase types, and splits
	from \cite{flickrentitiesijcv} we identified four tasks: \emph{scene}, \emph{animals}, \emph{vehicles}, \emph{instruments}.
	In this approach the first task contains a set of visual concepts that can also be appear in
	future tasks.
	
	Some statistics on number of images and vocabulary size for each task are given
	in table~\ref{tab:data_split} for both datasets.
	See the supplementary material for a detailed breakdown of classes appearing in each task and more details on these dataset splits. 
	MS-COCO does not provide a test set, so we randomly selected
	half of the validation set images and used them for testing only. Since images
	have at least five captions, we used the first five captions for each image as
	the target.

	\bgroup
	\def\arraystretch{1.2}
	\begin{table}
		\begin{center}
			\begin{minipage}[b]{0.5\linewidth}
				\centering
				\small
				\setlength\tabcolsep{6pt}
				\resizebox{\textwidth}{!}{
					\begin{tabular}{|c||rrr||r|}
						\hline
						\textbf{Task} & \textbf{Train} & \textbf{Valid} & \textbf{Test} & \textbf{Vocab (words)} \\
						\hline \hline
						\textbf{transport}    & 14,266  & 3,431  & 3,431 & 3,116 \\
						\textbf{animals}      & 9,314   & 2,273  & 2,273 & 2,178 \\
						\textbf{sports}       & 10,077  & 2,384  & 2,384 & 1,967 \\
						\textbf{food}         & 7,814   & 1,890  & 1,890 & 2,235 \\
						\textbf{interior}     & 17,541  & 4,340  & 4,340 & 3,741 \\
						\hline \hline
						\textbf{total}        & 59,012  & 14,318 & 14,318 & 6,344 \\
						\hline
						
					\end{tabular}
				}
				\vspace{0.01cm}
				
				(a) MS-COCO task statistics
			\end{minipage}
			\begin{minipage}[b]{0.49\linewidth}
				\centering
				\small
				\setlength\tabcolsep{6pt}
				\resizebox{\textwidth}{!}{
					\begin{tabular}{|c||rrr||r|}
						\hline
						\textbf{Task} & \textbf{Train} & \textbf{Valid} & \textbf{Test} & \textbf{Vocab (words)} \\
						\hline
						\textbf{scene}        & 5,000  & 170  & 170 & 2,714 \\
						\textbf{animals}      & 3,312  & 107  & 113 & 1,631 \\
						\textbf{vehicles}     & 4,084  & 123  & 149 & 2,169 \\
						\textbf{instruments}  & 1,290  & 42   & 42  & 848 \\
						\hline \hline
						\textbf{total}        & 18,283  & 607 & 636 & 4,123 \\
						\hline
					\end{tabular}
				}
				\vspace{0.01cm}
				
				(b) Flickr30k task statistics
				\setlength\tabcolsep{6pt}
			\end{minipage}
		\end{center}
		\caption{Number of images and words per task for our MS-COCO and Flickr30K splits.}
		\label{tab:data_split}
	\end{table}
	\egroup
	\subsection{Ablation study}
	We conducted a preliminary study on our split of MS-COCO to evaluate the impact
	of our proposed Recurrent Attention to Transient Tasks (RATT) approach. In this experiment
	we progressively introduce the attention masks described in
	section~\ref{sec:RAT}. We start with the basic captioning model with no
	forgetting mitigation, and so is equivalent to \emph{fine-tuning}. Then we
	introduce the mask on hidden state $h_n$ of the LSTM (along with the
	corresponding backward mask), and then the constant binary mask on the
	classifier that depends on the words of the current task, then the visual and
	word embedding masks, and finally the combination of all masks.
	
	\begin{figure}[b]
		\includegraphics[width=1.0\linewidth]{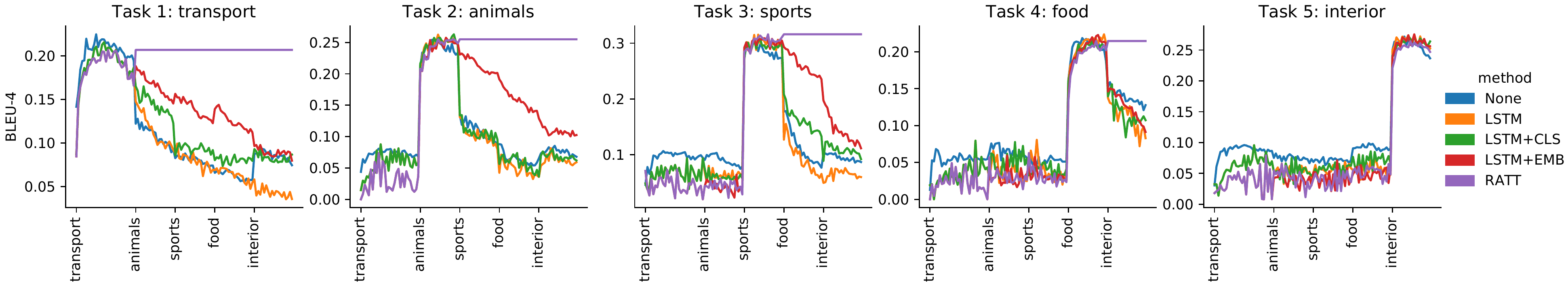}
		\caption{
			\label{fig:ablation_mscoco}
			BLEU-4 performance for several ablations at each epoch over the whole sequence of MS-COCO tasks.
		}
	\end{figure}
	
\begin{figure}
		\includegraphics[width=1.0\linewidth]{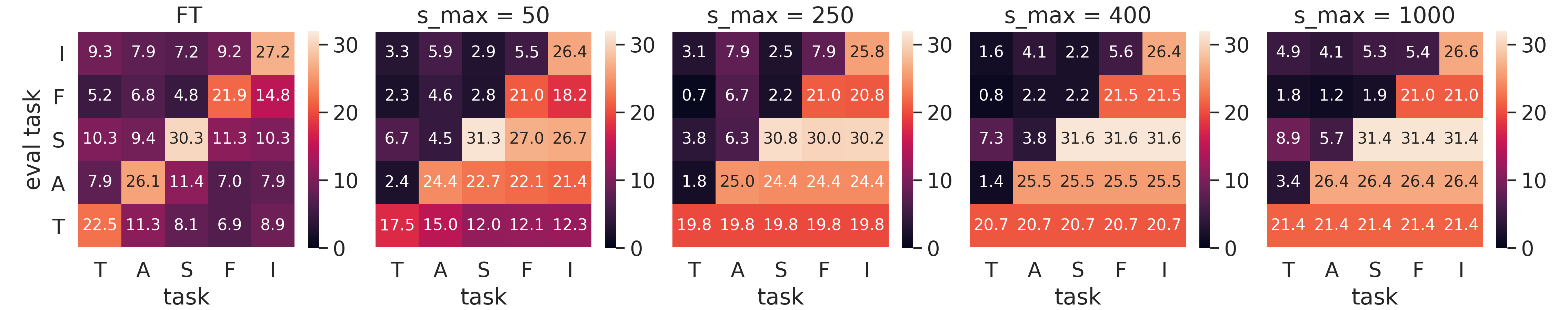}
		\vspace{-0.2in}
		\caption{
			\label{fig:coco_ratt_smax_heatmap}
			RATT ablation on the MS-COCO validation set using different $s_{max}$ values and fine-tuning baseline.
			Each heatmap reports BLEU-4 performance for one of the ablated models evaluated on different tasks
			at the end of the training of each task.
		}
	\end{figure}

	In figure~\ref{fig:ablation_mscoco} we plot the BLEU-4 performance of these
	configurations for each training epoch and each of the five MS-COCO tasks. Note
	that for later tasks the performance on early epochs (i.e. \emph{before}
	encountering the task) is noisy as expected -- we are evaluating performance on
	\emph{future} tasks. These results clearly show that applying the mask to LSTM
	decreases forgetting in the early epochs when learning a new task. However,
	performance continues to decrease and in some tasks the result is similar to
	fine-tuning. Even if the LSTM is forced to remember how to manage hidden states
	for previous tasks, the other parts of the network suffer from catastrophic
	forgetting. Adding the classifier mask improves the situation, but the main
	contribution comes from applying the mask to the embedding. Applying all masks
	we obtain zero or nearly-zero forgetting. 
	This depends on the $s_{max}$ value used during training: 
	in these experiments we use $s_{max}=400$, which results
	in zero forgetting of previous tasks.
	We also conducted an ablation study on the $s_{max}$ parameter. From the results in
	figure~\ref{fig:coco_ratt_smax_heatmap} we can see that higher $s_{max}$
	values improve old task performance, and sufficiently high values completely
	avoid forgetting. Using moderate values, however, can be helpful to increase
	performance in later tasks. See supplementary material for additional
	ablations.
	
	\subsection{Results on MS-COCO}
\begin{figure}
		\includegraphics[width=1\linewidth]{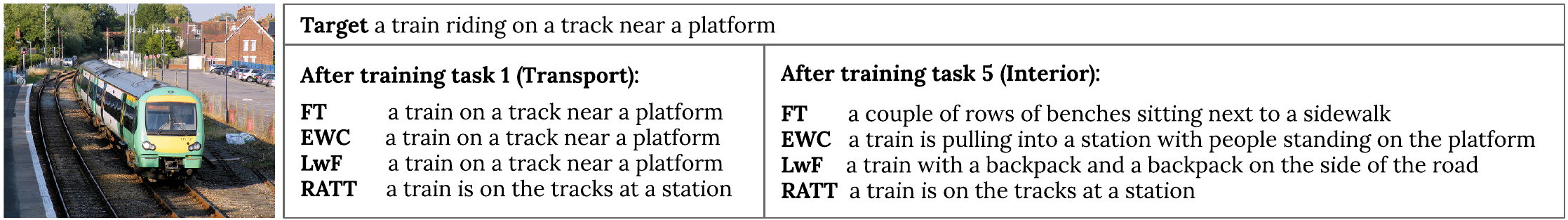}
		\caption{
			Qualitative results for image captioning on MS-COCO. Forgetting for baseline methods can be clearly observed. More results are provided in Supplementary Material.}
		\label{fig:qual_mscoco}
	\end{figure}

	In table~\ref{tab:coco_results} we report the performance of a fine-tuning
	baseline with no forgetting mitigation (FT), EWC, LwF, and RATT on our splits for
	the MS-COCO captioning dataset. The forgetting percentage is computed by taking
	the BLEU-4 score for each model after training on the last task and dividing it
	by the BLEU-4 score at the end of the training of each individual task. From the
	results we see that all techniques consistently improve
	performance on previous tasks when compared to the FT baseline.
	Despite the simplicity of EWC, the improvement over fine-tuning is
	clear, but it struggles to learn a good model for the last task. LwF instead
	shows the opposite behavior: it is more capable of learning the last task, but
	forgetting is more noticeable. RATT  achieves \emph{zero}
	forgetting on MS-COCO, although at the cost of some performance on the final
	task. This is to be expected, though, as our approach deliberately and
	progressively limits network capacity to prevent forgetting of old tasks.
	Qualitative results on MS-COCO are provided in figure~\ref{fig:qual_mscoco}.

	\bgroup
	\centering
	\def\arraystretch{1.3} 
	\setlength\tabcolsep{2pt}
	\begin{table}[b]
		
		\resizebox{1.0\textwidth}{!}{
			\begin{tabular}{ |c|cccc|cccc|cccc|cccc|cccc| } 
				
				\hline
				& \multicolumn{4}{c|}{\textbf{Transport}} & \multicolumn{4}{c|}{\textbf{Animals}} & \multicolumn{4}{c|}{\textbf{Sports}} & \multicolumn{4}{c|}{\textbf{Food}} & \multicolumn{4}{c|}{\textbf{Interior}} \\
				& \textbf{FT}	& \textbf{EWC}	& \textbf{LwF}	& \textbf{RATT}	& \textbf{FT}	& \textbf{EWC}	& \textbf{LwF}	& \textbf{RATT}	& \textbf{FT}	& \textbf{EWC}	& \textbf{LwF}	& \textbf{RATT}	& \textbf{FT}	& \textbf{EWC}	& \textbf{LwF}	& \textbf{RATT}	& \textbf{FT}	& \textbf{EWC}	& \textbf{LwF}	& \textbf{RATT}\\
				\hline
				%BLEU-1	& .5285	& .6322	& .5788	& .6969	& .4939	& .6044	& .5486	& .6958	& .5756	& .7023	& .6313	& .7726	& .6043	& .6010	& .6529	& .6966	& .7180	& .6647	& .7095	& .7159\\
				\textbf{BLEU-4}	
				& .0928	& .1559	& .1277	& \textbf{.2126}
				& .0816	& .1545	& .1050	& \textbf{.2468	}
				& .0980 & .2182	& .1491	& \textbf{.3161}
				& .1510	& .1416	& .1623 & \textbf{.2169}
				& .2712	& .2107	& .2537	& \textbf{.2727}
				\\
				\textbf{METEOR}	
				& .1472	& .1919	& .1708	& \textbf{.2169}
				& .1396	& .1779	& .1577	& \textbf{.2349	}
				& .1639	& .2209	& .1918	& \textbf{.2707}
				& .1768	& .1597	& .1962	& \textbf{.2110	}
				& \textbf{.2351}	& .1967	& .2286	& .2257
				\\
				%ROUGE$_L$	& .3882	& .4642	& .4280	& .5078	& .3751	& .4569	& .4142	& .5285	& .4307	& .5101	& .4746	& .5766	& .4386	& .4360	& .4661	& .5039	& .5381	& .5014	& .5291	& .5376\\
				\textbf{CIDEr}	
				& .2067	& .4273	& .3187	& \textbf{.6349}
				& .1480	& .4043	& .2158	& \textbf{.7249	}
				& .1680	& .5146	& .3277	& \textbf{.8085}
				& .2668	& .2523	& .3816	& \textbf{.5195	}
				& \textbf{.6979}	& .4878	& .6554	& .6536\\
				\hline
				\hline
				\textbf{\% forgetting} & 59.1	& 31.2	& 43.7	& 0.0	& 67.5	& 33.8	& 45.0	& 0.0	& 68.9	& 23.6	& 45.0	& 0.0	& 32.8	& 14.6	& 16.5	& 0.0	& N/A	& N/A	& N/A	& N/A \\
				\hline
			\end{tabular}
		}
		\setlength\tabcolsep{6pt}
		\caption{Performance on MS-COCO. Numbers are the per-task performance after
			training on the \emph{last} task. Per-task forgetting in the last row is the
			BLEU-4 performance after the last task divided by the BLEU-4 performance
			measured immediately after learning each task.}
		\label{tab:coco_results}
	\end{table}
	\egroup

	\subsection{Results on Flickr30k}
	In table~\ref{tab:flickr_results} we report performance of a fine-tuning
	baseline with no forgetting mitigation (FT), EWC, LwF, and RATT on our Flickr30k
	task splits. Because these splits are based on \emph{incremental visual
		categories}, it does not reflect a classical continual-learning setup that
	enforce disjoint categories to maximize catastrophic forgetting: not only there
	are common words that share the same meaning between different tasks, but some
	of the visual categories in early tasks are also present in future ones. For
	this reason, learning how to describe task $t=1$ also implies learning at least
	how to partially describe future tasks, so forward and backward transfer is
	significant.
	
	Despite this, we see that all approaches increase performance on old
	tasks (when compared to FT) while retaining good performance on the last one. Note that both RATT and
	LwF result in \emph{negative forgetting}: in these cases the training of a new
	task results in backward transfer that increases performance on an old one. EWC
	improvement is marginal, and LwF behaves a bit better and seems more capable
	of exploiting backward transfer. RATT backward transfer is instead limited by the
	choice of a high $s_{max}$, which however guarantees nearly zero forgetting. 

	\bgroup
	\centering
	\def\arraystretch{1.3}
	\setlength\tabcolsep{3pt}
	\begin{table}
		\begin{center}
			\resizebox{1\textwidth}{!}{
				\begin{tabular}{|c|cccc|cccc|cccc|cccc|} 
					\hline
					&  \multicolumn{4}{c|}{\textbf{Scene}} & \multicolumn{4}{c|}{\textbf{Animals}} & \multicolumn{4}{c|}{\textbf{Vehicles}} & \multicolumn{4}{c|}{\textbf{Instruments}} \\
					& \textbf{FT}	& \textbf{EWC}	& \textbf{LwF}	& \textbf{RATT}	& \textbf{FT}	& \textbf{EWC}	& \textbf{LwF}	& \textbf{RATT}	& \textbf{FT}	& \textbf{EWC}	& \textbf{LwF}	& \textbf{RATT}	& \textbf{FT}	& \textbf{EWC}	& \textbf{LwF}	& \textbf{RATT}\\
					\hline
					%BLEU-1 &	.5007 &	.5665 &	.5867 &	.5728 &	.5192 &	.5707 &	.5773 &	.6164 &	.5199 &	.5736 &	.5793 &	.5843 &	.5984 &	.6845 &	.6498 &	.6554\\
					\textbf{BLEU-4}
					&	.1074 &	.1370 &	.1504 &	\textbf{.1548} 
					&	.1255 &	.1381 &	.1384 &	\textbf{.1921} 
					&	.1083 &	.1332 &	.1450 &	\textbf{.1724} 
					&	.1909 &	.2313 &	.1862 &	\textbf{.2386} \\
					\textbf{METEOR}
					&	.1570 &	.1722 &\textbf{.1851} &	.1710 
					&	.2046 &	.1833 &	.1954 &	\textbf{.2107} 
					&	.1625 &	.1770 &	\textbf{.1847} &	.1750 
					&	\textbf{.1933} &	.1714 &	.1876 &	.1782 \\
					%ROUGE$_L$ &	.3606 &	.3864 &	.4022 &	.3972 &	.3956 &	.4195 &	.4063 &	.4780 &	.3745 &	.3850 &	.3918 &	.4223 &	.4608 &	.4779 &	.4615 &	.4922\\
					\textbf{CIDEr}
					&	.1222 &	.1688 &	.2402 &	\textbf{.2766 }
					&	.2460 &	.2755 &	.2756 &	\textbf{.4708 }
					&	.1586 &	.1315 &	.1748 &	\textbf{.2988 }
					&	.2525 &	.2611 &	\textbf{.2822} &	.2329 \\
					\hline
					\hline
					\textbf{\% forgetting} 
					&	31.1 &	11.3 &	2.7 &	-2.5 
					&	38.7 &	19.2 &	-15.1 &	0.0 
					&	35.6 &	4.9 &	-1.5 &	0.0 
					&	N/A &	N/A &	N/A &	N/A\\
					\hline
				\end{tabular}
			}
			\setlength\tabcolsep{6pt}
		\end{center}
		\caption{Performance on Flickr30K. Evaluation is the same as for MS-COCO.}
		\label{tab:flickr_results}

	\end{table}
	\egroup
	
	\subsection{Human evaluation experiments}
	We performed an evaluation based on human quality judgments 
	using 200 images (40 from each task) from the MS-COCO test splits.
	We generated captions with RATT, EWC, and LwF after training on the
	last task and then presented ten users with an image and RATT and baseline
	captions in random order. Users were asked (using forced choice) to select which
	caption best represents the image content. A similar evaluation was performed for the Flickr30k dataset with twelve users. The percentage of users who chose RATT over the baseline are  given in the table \ref{table:human-evaluation}.
	These results on MS-COCO dataset confirm that RATT is superior on all tasks, while on Flickr30k there is some uncertainty on the first task, especially when comparing RATT with LwF. Note that for the last task of each dataset there is no forgetting, so it is expected that baselines
	and RATT perform similarly.

	\bgroup
	\def\arraystretch{1.2}
	\begin{table}
		\begin{center}
			\resizebox{1\textwidth}{!}{
				\begin{tabular}{|l|ccccc|cccc|}
					\hline
					\multicolumn{1}{|c|}{} & 
					\multicolumn{5}{c|}{\textbf{MS-COCO}} &  
					\multicolumn{4}{c|}{\textbf{Flickr30k}}\\
					
					& \textbf{T} & \textbf{A} & \textbf{S} & \textbf{F} & \textbf{I} &  \textbf{S} & \textbf{A} & \textbf{V} & \textbf{I} \\ 
					\hline
					\textbf{RATT vs EWC} & 75.0\% & 77.5\% & 72.5\% & 85.0\% & 57.5\% &
					61.8\% & 76.4\% & 67.3\% & 59.5\% \\
					\hline
					\textbf{RATT vs LwF} & 77.5\% & 82.5\% & 75.0\% & 62.5\% & 47.5\%& 
					45.5\% & 69.1\% & 63.6\% & 59.5\% \\
					\hline
				\end{tabular}
			}
			\vspace{0.15cm}
			\caption{Human captioning evaluation on both MS-COCO and Flickr30k. 
				For each task, we report the percentage of examples for which users preferred the caption generated by RATT.}
			\label{table:human-evaluation}\vspace{-0.15in}
		\end{center}
	\end{table}
	\egroup

	%%%%%%%%%%%%%%%%%%%%%%%%%%%%%%%%%%%%%%%%%%%%%%%%%%%%%%%%%%%%%%%%%%%%%%%%%%%%%%%%%%%%%%%%%%%%%%%
	\section{Conclusions}\label{sec:conclusions}
	In this paper we proposed a technique for continual learning of image captioning
	networks based on Recurrent Attention to Transient Tasks (RATT). Our approach is
	motivated by a feature of image captioning not shared with other continual
	learning problems: tasks are composed of \emph{transient} classes (words) that
	can be shared across tasks. We also showed how to adapt Elastic Weight
	Consolidation and Learning without Forgetting, two representative approaches of
	continual learning, to the recurrent image captioning networks. We proposed task
	splits for the MS-COCO and Flickr30k image captioning datasets, and our
	experimental evaluation confirms the need of recurrent task attention in order to
	mitigate forgetting in continual learning with sequential, transient tasks.
	RATT is capable of zero forgetting at the expense of plasticity and backward transfer: the ability to adapt to new tasks is limited by the number of free neurons and it is difficult to exploit knowledge from future tasks to better predict older ones. 
	The focus of this work is on how a simple encoder-decoder image captioning model forget, which limits the quality of captions when comparing with current state-of-the-art. As future work, we are interested in applying the developed method in more complex captioning systems.

	%%%%%%%%%%%%%%%%%%%%%%%%%%%%%%%%%%%%%%%%%%%%%%%%%%%%%%%%%%%%%%%%%%%%%%%%%%%%%%%%%%%%%%%%%%%%%%%
	
	\section*{Broader impact}
	Automatic image captioning has applications in image indexing, Content-Based
	Image Retrieval (CBIR), and industries like commerce, education, digital
	libraries, and web searching. Social media platforms could use it to directly
	generate descriptions of multimedia content. Image captioning systems can
	support peoples with disabilities, providing an access to the visual content
	unreachable before by changing its representation. Continual learning contrasts
	with the joint-training paradigm in common use today. It has the advantage that it can better protect privacy concerns since,
	once learned, the data not need to be retained. Furthermore, it is more 
	efficient since networks continue learning and are not initialized from scratch 
	every time a new task arrives.
	This ability is crucial for development of systems like virtual
	personal assistants, where the adaption to new tasks and environments is fundamental,
	especially where multi-modal communication channels are ubiquitous.
	Finally, the algorithm considered in this paper will reflect the biases present
	in the dataset. Therefore, special care should be taken when applying this
	technique to applications where possible biases in the dataset might result in
	biased outcomes towards minority and/or under-represented groups in the data.
	
	\begin{ack}
		We acknowledge the support from Huawei Kirin Solution.  We thank NVIDIA Corporation for donating the Titan XP GPU that was used to conduct the experiments. We also acknowledge the project PID2019-104174GB-I00 of Ministry of Science of Spain and the ARS01\_00421 Project "PON IDEHA - Innovazioni per l'elaborazione dei dati nel settore del Patrimonio Culturale" of the Italian Ministry of Education. Our acknowledged partners funded this project.
	\end{ack}

	\bibliography{ratt}
	\bibliographystyle{plain}
	
\end{document}

\maketitle

	\section{Task splits for incremental captioning}
	Here we first describe the two splitting procedures we propose that are
	applicable to captioning datasets with categorical annotations. Then we describe
	how we apply them to the MS-COCO~\cite{lin2014mscoco} and
	Flick30k~\cite{flickrentitiesijcv} datasets.

	\subsection{Disjoint visual categories}
	
	We exploit categorical image annotations available in many captioning datasets.
	If each image in the dataset belongs to a single category, we can simply define
	each task as a set of categories that does not overlap with any other task. If
	an image can belong to multiple categories we instead use the following
	procedure:
	\begin{enumerate}
		\item \textbf{Define $K$ tasks}. Tasks are sets $\mathcal{C}_t$ of categories such that: $\mathcal{C}_i \cap \mathcal{C}_j = \emptyset \ \forall i \ne j$.
		\item \textbf{Identify candidate example sets}. For each task $t$ select all the examples in
		the original dataset having at least one of the labels in common with task $t$ categories:
		\begin{equation}
		\mathrm{P}_t = \{i \ | \ \exists \ c \in \mathcal{C}_t \ \text{s.t.} \ y_c^i = 1 \}  
		\end{equation}
		where $i$ is the index of example in the original dataset and $y^i \in \{0, 1\}^{|\mathcal{C}_t|}$ is a multi-label vector such that $y^i_c = 1  \Leftrightarrow \mbox{the $i$-th example belongs to category $c$}$.
		
		\item \textbf{Identify common examples sets}. Find common examples in candidate sets: 
		$\mathrm{Q}_{i,j} = 
		\mathrm{P}_i \cap \mathrm{P}_j$
		
		\item \textbf{Define final task examples}. Define example sets of each task $t$ as: 
		$\mathrm{E}_t = \mathrm{P}_t \setminus \cup_{i \ne t}(\mathrm{Q}_{t,i})$
	\end{enumerate}
	
	This guarantees that if an image belongs to multiple tasks due to its labels, it
	will be completely pruned from the dataset instead of added to both or added to
	only one.
	
	\subsection{Incremental visual categories} As an alternative to
	visually-disjoint task splits, we also evaluate continual image captioning in a more real-life setting, where a first task contains a set of
	visual concepts that can reappear in following tasks. Subsequent tasks contain new or more specific concepts, without the guarantee of having no
	overlap with the already seen data. 
	The idea is to train the network over general
	concepts and then progressively train it on more specific ones. The network
	should continue to perform well on old tasks without overfitting to the more
	recently seen. The procedure is as follows (note that two first steps are the same as before):
	\begin{enumerate}
		\item \textbf{Define $K$ tasks}. Tasks are sets $\mathcal{C}_t$ of categories. 
		\item \textbf{Identify candidate example sets}. As in point (2) of the previous procedure:
		\begin{equation}
		\mathrm{P}_t = \{i \ | \ \exists \ c \in \mathcal{C}_t \ \text{s.t.} \ y_c^i = 1 \}
		\end{equation}
		where $i$ is the index of an example in the original dataset and $y^i \in \{0, 1\}^{|\mathcal{C}_t|}$
		is a multi-label vector such that $y^i_c = 1  \Leftrightarrow \mbox{the $i$-th example belongs to category $c$}$.
		
		\item \textbf{Define final task examples}.  Define example sets of each task $t$ as:
		$\mathrm{E}_t = \mathrm{P}_t \setminus \cup_{i=t}^{K}( \mathrm{P}_t \cap \mathrm{P}_i )$
	\end{enumerate}
	
	Given the sets $E_t$ we define the training set for the task $t$ as:
	\begin{equation}
	\mathcal{D}_t = \{x^i, S^{i,1}, S^{i,2}, ... S^{i,\Sigma} \ | \ i \in \mathrm{E}_t\}
	\end{equation}
	where $S^{i,j}$ is a sentence describing image $x^i$ and $\Sigma$ is the number of
	sentences describing each image.

	\subsection{An MSCOCO task split}
	We applied the \emph{disjoint visual categories} splitting procedure to arrive
	at the following task split for MS-COCO~\cite{lin2014mscoco}:
	\begin{itemize}
		\itemsep0em 
		\item \textbf{transport}: bicycle, car, motorcycle, airplane, bus, train, truck, boat.
		\item \textbf{animals}: bird, horse, sheep, cow, elephant, bear, zebra, giraffe.
		\item \textbf{sports}: snowboard, sports ball, kite, baseball bat, baseball~glove, skateboard, surfboard, tennis~racket.
		\item \textbf{food}: banana, apple, sandwich, orange, broccoli, carrot, hot~dog, pizza, donut, cake.
		\item \textbf{interior}: chair, couch, potted~plant, bed, toilet, tv, laptop, mouse, remote, keyboard, cell~phone, microwave, oven, toaster, sink, refrigerator.
	\end{itemize}
	We removed categories \emph{dog} and \emph{cat} from \emph{animals} because
	objects of these classes are very likely to appear also in images of
	\emph{interiors} and \emph{sports} tasks. For the same reason we removed
	\emph{dinning table} from \emph{interior} because of the likely overlap with the
	\emph{food} task.
	In table \ref{tab:overlap} we report word overlaps between tasks for our MS-COCO splits.
	From this breakdown we see that the task vocabularies are approximately the same size (between around 2,000 and 3,000 words), and there is significant 
	overlap between all tasks.
	
	\bgroup
	
	\def\arraystretch{1.2}
	\begin{table}
		\begin{center}
			\resizebox{0.8\textwidth}{!}{	
				\begin{tabular}{crrrrr}
					\hline
					& \myalign{c}{\textbf{T}} & \myalign{c}{\textbf{A}} & \myalign{c}{\textbf{S}} & \myalign{c}{\textbf{F}} &
					%			 \myalign{c|}{\textbf{I}} \\ 				
					\myalign{c}{\textbf{I}} \\ 				
					\hline
					\textbf{T} & 3,116 (100.0\%)	& 1,499 (48.11\%) & 1,400 (44.93\%) & 1,222 (39.22\%) & 1,957 (62.80\%) \\ 
					%			\hline
					\textbf{A} & 1,499 (48.11\%)	& 2,178 (100.0\%) & 1,175 (53.95\%) & 1,025 (47.06\%) & 1,492 (68.50\%) \\ 
					%			\hline
					\textbf{S} & 1,400 (44.93\%)	& 1,175 (53.95\%) & 1,967 (100.0\%) & 933   (47.43\%) & 1,355 (68.89\%) \\ 
					%			\hline
					\textbf{F} & 1,222 (39.22\%)	& 1,025 (47.06\%) &  933  (47.43\%) & 2,235 (100.0\%) & 1,530 (68.46\%) \\ 
					%			\hline
					\textbf{I} & 1,957 (62.80\%)	& 1,492 (68.50\%) & 1,355 (68.89\%) & 1,530 (68.46\%) & 3,741 (100.0\%) \\ 
					\hline
			\end{tabular}}
		\end{center}
		\vspace{0.2cm}
		\caption{word overlaps between tasks for our MS-COCO splits.}
		\label{tab:overlap}
		
	\end{table}
	\egroup
	
	\subsection{A Flickr30k task split}
	
\begin{figure}
		\centering
		\includegraphics[width=0.45\linewidth]{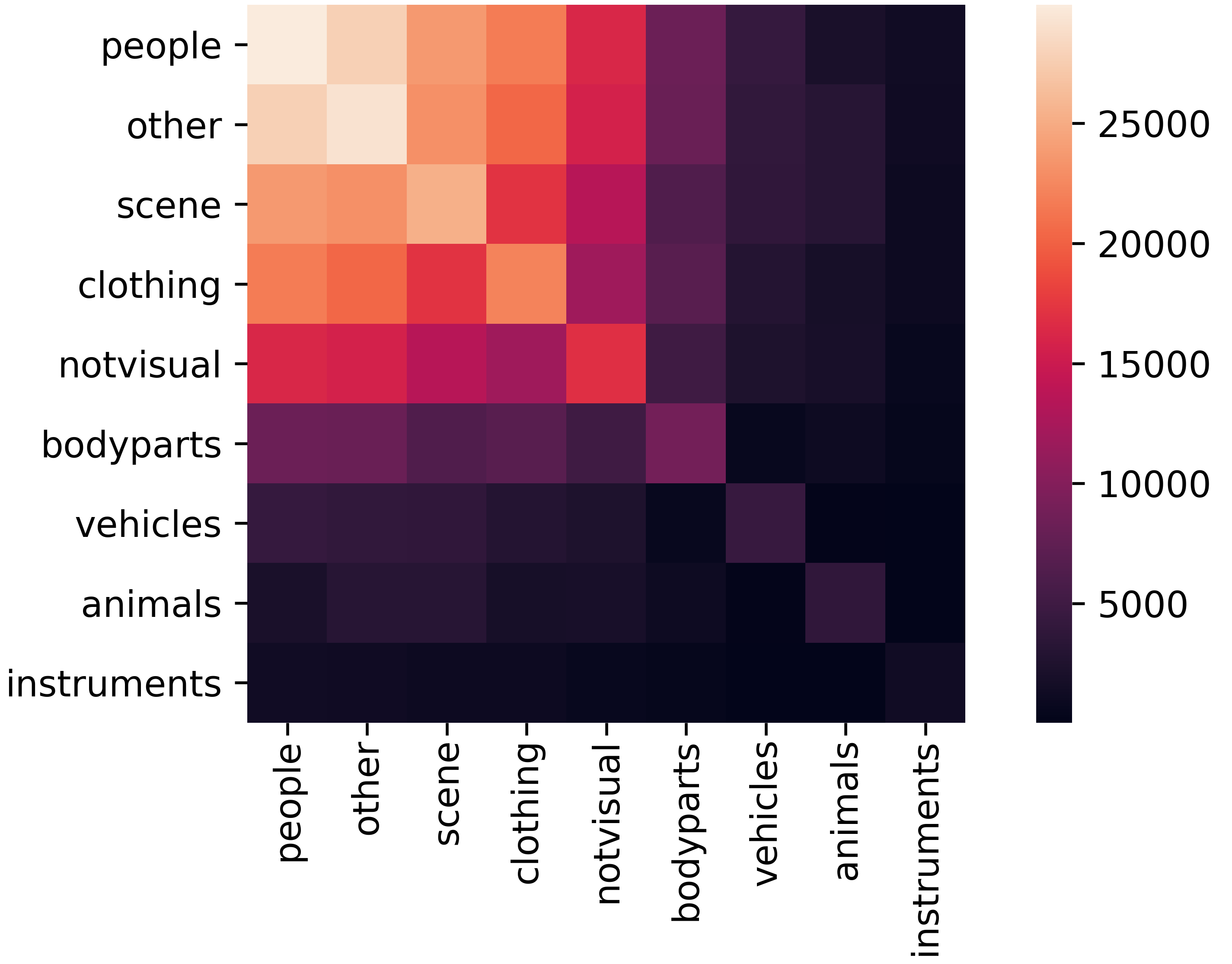}
		\vspace{-0.2cm}
		\caption{Flicker30r co-occurrence matrix for assigned categories.}
		\label{fig:flickr-phrases}
	\end{figure}

	In the Flickr30k Entities~\cite{flickrentitiesijcv} dataset we have five
	captions per image and each caption is labeled with a set of \emph{phrase types}
	that refers to parts of the sentence. We use the union of all phrase types
	associated to each example as the set of categories for that example. A subset
	of these categories is used to split the dataset using the \emph{incremental
		visual categories} procedure. For this dataset we use a single category per task,
	so tasks are named after assigned categories. The list of categories (tasks) is:
	\textbf{scene}, \textbf{animals}, \textbf{vehicles}, and \textbf{instruments}.
	If a category is over-represented, random sub-sampling is done to get maximum of
	7,500 examples (like in the case of \textbf{scene}). Moreover, the most
	common phrase type is \textbf{people} and we omit it in purpose because almost
	all photos contain people. In figure~\ref{fig:flickr-phrases} we give the
	co-occurrence matrix between Flickr30k images and categories based on phrases
	types from~\cite{flickrentitiesijcv}. The influence of the \textbf{people}
	category is clearly visible. 
	
	\section{Additional training details}
	\label{sec:train_details}
	In this section give additional details about network training for the
	approaches we implemented to prevent catastrophic interference in LSTM
	captioning networks.
	
	\subsection{Weight Regularization}
	Optimizing $\mathcal{L}^t_{EWC}(\theta)$ (equation~(13) in the main paper) we
	obtain $\hat{\theta}^t$ and when proceeding to the task $t+1$ we again use
	$\mathcal{L}^t_{EWC}$ but supplying instead $\hat{\theta}^t_i$ as its argument.
	$\mathcal{L}^t_{EWC}$ is applied to every trainable weight of our network, but
	we have special cases for the weights of the word embedding and final
	classifier: these are only \emph{partially} shared between tasks. At each new
	task some of the weights will be completely new since they are related to new
	words, we do not want to force these weights to stay where they are since they
	have never been trained before, and so we do not regularize them. Note this
	problem is not present in the standard continual learning for classification
	because each new task has a disjoint set of classes and a dedicated classifier
	is used per task.
	
	\subsection{Recurrent learning without forgetting}
	The final loss for training the decoder network with LwF is:
	\begin{equation}
	\label{eq:lwc_total_loss}
	\mathcal{L}^{t}(x, S) =  
	\mathcal{L}(x, S) +  \mathcal{L}^{t}_{\mathrm{LwF}}(\hat{p}^t, p^{t-1}) =
	- \sum_{n=1}^N \left[ \log p_n(S_n) - \lambda H(\gamma(\hat{p}^t_n), \gamma(p^{t-1}_n)) 
	\right]
	\end{equation}
	where $\lambda$ is the hyperparameter weighting the importance of the previous task.  
	Note that differently from~\cite{li2017learning}, we do not fine-tune the classifier of the old 
	network because we use a single, incremental word classifier.
	
	\subsection{Recurrent Attention to Trainsent Tasks}
	The final loss for training the decoder network with RATT is:
	
	\begin{equation}
	\mathcal{L}^{t}(x, S) =  \mathcal{L}(x, S) + \mathcal{L}^t_{a} = 
	- \sum_{n=1}^N \log p_n(s_n)
	+ \lambda \frac { \sum_{i} a^t_{x, i} (1-a^{<t}_{x, i})}{ \sum_{i} (1-a^{<t}_{x, i})} 
	+ \lambda \frac{ \sum_{i} a^t_{h, i} (1-a^{<t}_{h, i})}{ \sum_{i} (1-a^{<t}_{h, i})}.
	\end{equation}
	where $\lambda$ is the hyperparameter weighting the importance of future tasks: for larger
	$\lambda$, fewer neurons will be allocated to the current task (and more neurons will be
	available for the future tasks).
	
	The backpropagation updates for each LSTM gate matrix are:
	
	\begin{tabular}{cc}
		\begin{minipage}[t]{0.45\textwidth}
			\begin{eqnarray}
			W_{ih} &\leftarrow& W_{ih} - \lambda B^t_{ih} \odot \frac{\partial \mathcal{L}^t}{\partial W_{ih}} \\
			W_{ix} &\leftarrow& W_{ix} - \lambda B^t_{ix} \odot \frac{\partial \mathcal{L}^t}{\partial W_{ix}} \\
			W_{oh} &\leftarrow& W_{oh} - \lambda B^t_{oh} \odot \frac{\partial \mathcal{L}^t}{\partial W_{oh}} \\
			W_{ox} &\leftarrow& W_{ox} - \lambda B^t_{ox} \odot \frac{\partial \mathcal{L}^t}{\partial W_{ox}}
			\end{eqnarray}
		\end{minipage}
		&
		\begin{minipage}[t]{0.45\textwidth}
			\begin{eqnarray}
			W_{fh} &\leftarrow& W_{fh} - \lambda B^t_{fh} \odot \frac{\partial \mathcal{L}^t}{\partial W_{fh}} \\
			W_{fx} &\leftarrow& W_{fx} - \lambda B^t_{fx} \odot \frac{\partial \mathcal{L}^t}{\partial W_{fx}} \\
			W_{gh} &\leftarrow& W_{gh} - \lambda B^t_{gh} \odot \frac{\partial \mathcal{L}^t}{\partial W_{gh}} \\
			W_{gx} &\leftarrow& W_{gx} - \lambda B^t_{gx} \odot \frac{\partial \mathcal{L}^t}{\partial W_{gx}}
			\end{eqnarray}
		\end{minipage}
	\end{tabular}
	
	During training, we applied the gradient compensation procedure described 
	in~\cite{serra2018overcoming} to help training the task-embedding matrices $A_x$ and $A_h$:
	\begin{eqnarray}
	A_{x,i} &\leftarrow& \frac{s_{max} [ \mathrm{cosh}(s A_{x,i}t^T) +1 ] }{s [ \mathrm{cosh}(A_{x,i}t^T) +1 ] } \frac{\partial \mathcal{L}^t}{\partial A_{x,i}} \\
	A_{h,i} &\leftarrow& \frac{s_{max}  [ \mathrm{cosh}(s A_{h,i}t^T ) +1 ] }{s [ \mathrm{cosh}(A_{h,i}t^T) +1 ] } \frac{\partial \mathcal{L}^t}{\partial A_{h,i}}.
	\end{eqnarray}
	Moreover, for numerical stability, we clamp $|s \ A_{x,i}t^T| \leq 50$ and
	$|s \ A_{h,i}t^T| \leq 50$.

	\section{Additional ablations}
	\label{sec:ablations}
\begin{figure}[H]
		\includegraphics[width=1.0\linewidth]{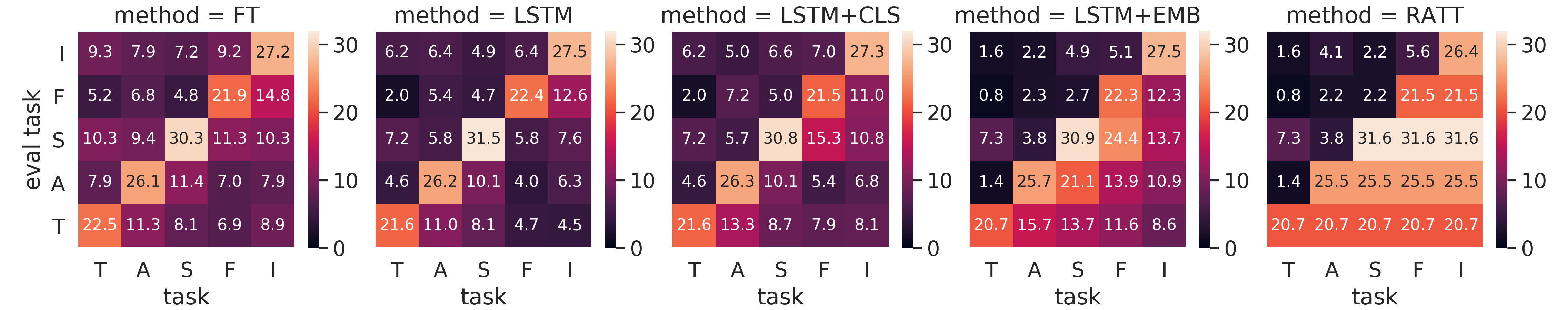}
		\vspace{-0.5cm}
		\caption{
			RATT ablation on the MS-COCO validation set using different attention masks.
			Each heatmap report BLEU-4 performance for one of the ablated models
			evaluated on different tasks at the end of the training of each task. }
		\label{fig:ratt_ablation_heatmaps}
	\end{figure}

	In figure~\ref{fig:ratt_ablation_heatmaps} we provide a different visualization
	of the RATT ablation reported in the main paper where we apply attention masking
	in different layers of the decoder architecture. In
	figure~\ref{fig:ratt_ablation_heatmaps} we observe an increase of performance on
	old tasks when the classifier mask is used, and even more clearly when the
	embedding mask is used. Even further improvement in the performance is made when all the attention masks (the RATT approach) are used and there is no forgetting.
	
\begin{figure}[H]
		\begin{center}
			\includegraphics[width=1.0\linewidth]{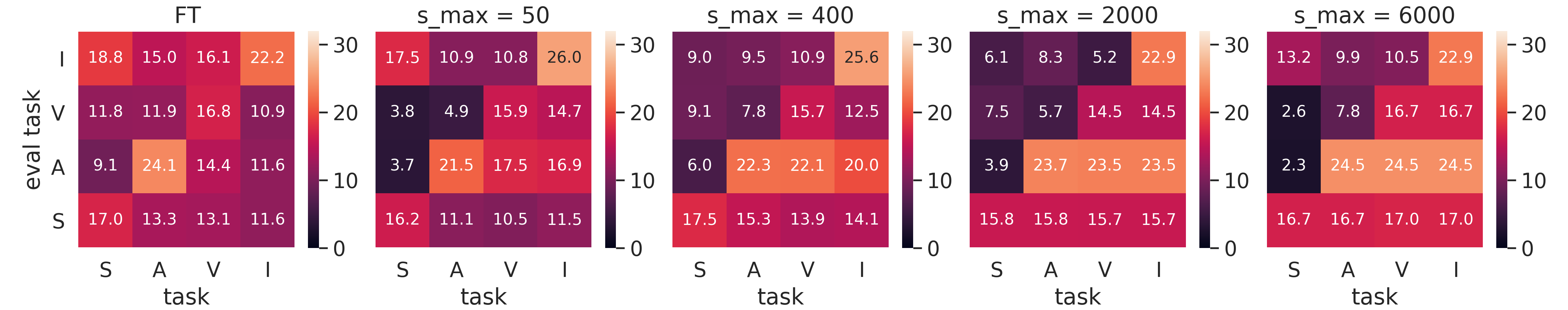}
			\vspace{-0.5cm}
			\caption{
				\label{fig:flickr_ratt_smax_heatmap}
				RATT ablation on Flickr30k validation set using different $s_{max}$ values and finetuning baseline.
				Each heatmap reports BLEU-4 performance for one of the ablated models evaluated on different tasks
				at the end of the training of each task.
			}
		\end{center}
	\end{figure}

\begin{figure}[H]
		\includegraphics[width=1.0\linewidth]{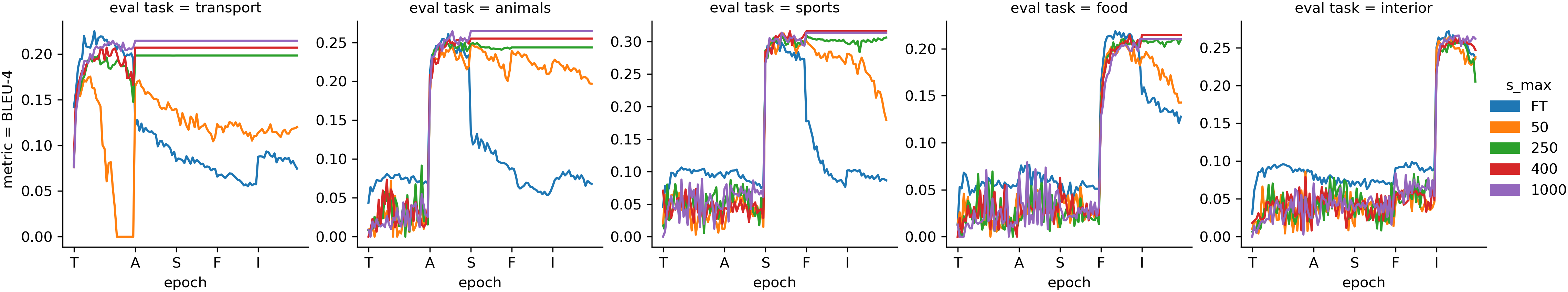}
		\vspace{-0.5cm}
		\caption{
			\label{fig:coco_ratt_smax_all_epochs}
			RATT ablation on the MS-COCO validation set using different $s_{max}$ values
			and finetuning baseline. Each plot reports BLEU-4 performance evaluated on
			one of the tasks at different training epochs and different training tasks
			for each of the ablated models.}
		
	\end{figure}

\begin{figure}[H]
		\begin{center}
			\includegraphics[width=0.9\linewidth]{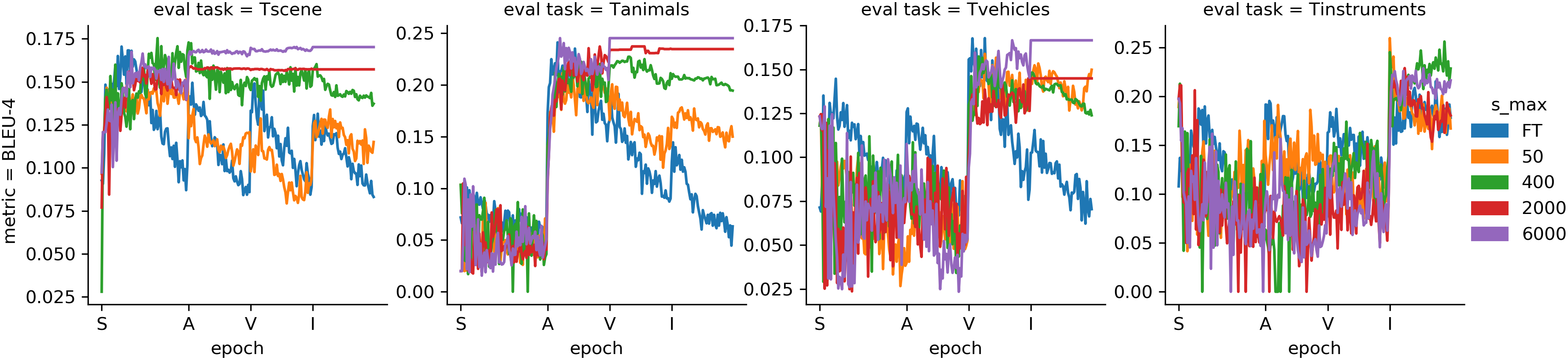}
			\vspace{-0.05cm}
			\caption{
				\label{fig:flickr_ratt_smax_all_epochs}
				RATT ablation on Flickr30k validation set using different $s_{max}$ values and finetuning baseline.
				Evaluation is the same as MS-COCO (figure \ref{fig:coco_ratt_smax_all_epochs}).
			}
			
		\end{center}
	\end{figure}

	We also conducted an ablation study on the $s_{max}$ parameter on Flickr30k, and  
	results are reported in figure~\ref{fig:flickr_ratt_smax_heatmap}.
	Different visualizations for this ablation are shown in figure~\ref{fig:coco_ratt_smax_all_epochs} 
	(for MS-COCO) 
	and \ref{fig:flickr_ratt_smax_all_epochs} (for Flickr30k).
	From the MS-COCO experiment backward transfer for RATT is not noticeable,
	while for the Flickr30k case we observe in figure~\ref{fig:flickr_ratt_smax_all_epochs} 
	that lower $s_{max}$ values result in a small boost in performance for previous 
	tasks when the training is started on each new one.
	However at the end of each training session the forgetting is always greater 
	than the backwards transfer. 
	Moreover, the model with highest $s_{max}$ (purple line in figure~\ref{fig:flickr_ratt_smax_all_epochs}) still shows a small
	amount of backward transfer, and in this case the performance gain is retained
	until the end of training.
	This is also noticeable in the last heatmap of figure~\ref{fig:flickr_ratt_smax_heatmap}
	for the first task (Sport) (bottom row).


	%\clearpage
	\section{Additional experimental analysis}
	\label{sec:analysis}
	In this section we give additional comparative performance analysis for RATT, EWC, and LwF on both datasets. 
	
	\subsection{Learning and forgetting on MS-COCO}
\begin{figure}[H]
		\includegraphics[width=1.0\linewidth]{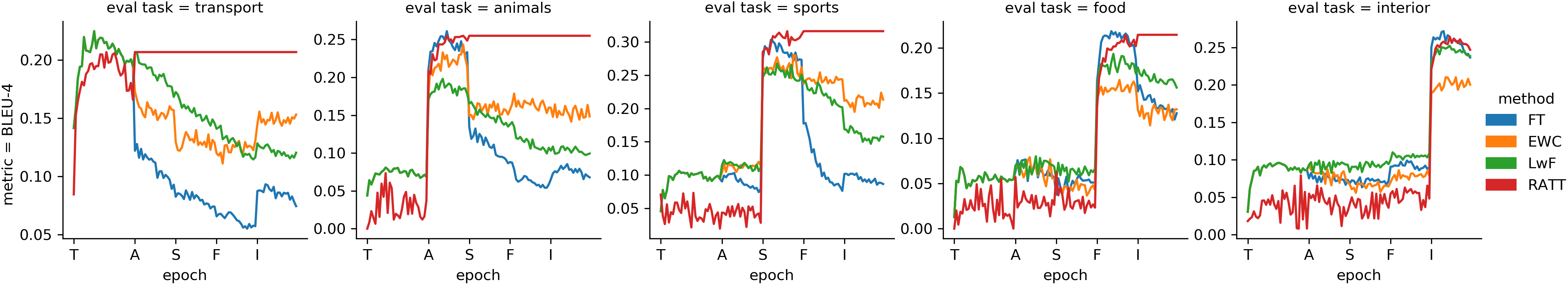}
		\vspace{-0.5cm}
		\caption{
			\label{fig:coco_comparison_epochs}
			Comparison for all approaches on MS-COCO validation set. 
			Each plot reports BLEU-4 performance evaluated on one of the tasks at
			different training epochs and different training tasks for each of the
			ablated models. }
		\vspace{-0.3cm}
	\end{figure}

\begin{figure}[H]
		\includegraphics[width=1.0\linewidth]{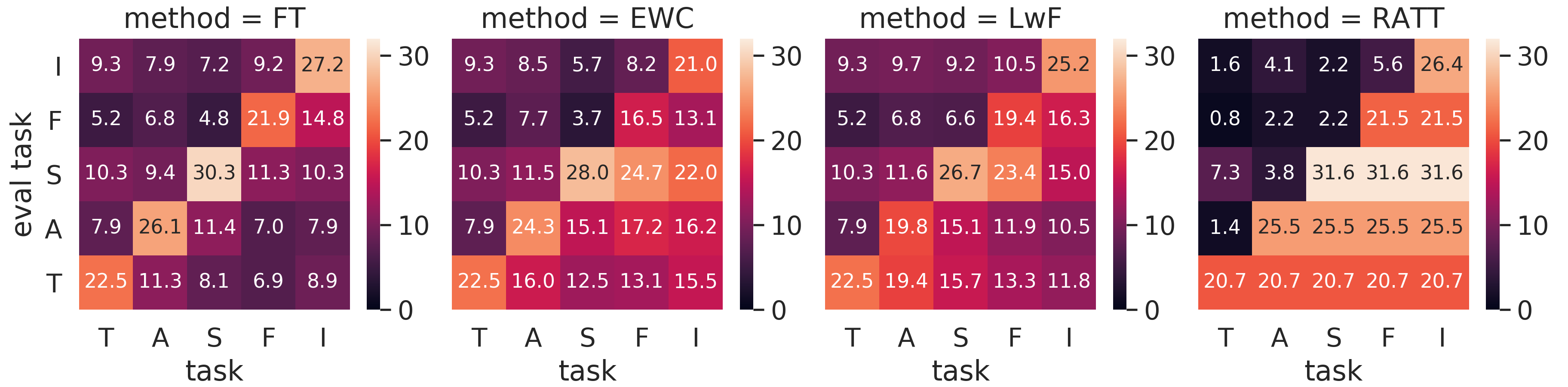}
		\vspace{-0.5cm}
		\caption{
			\label{fig:coco_comparison_heatmaps}
			Comparison for all approaches on MS-COCO validation set. 
			Each heatmap reports BLEU-4 performance for one of the models evaluated on different tasks
			at the end of the training of each task.
		}
		\vspace{-0.3cm}
	\end{figure}

	In figures \ref{fig:coco_comparison_epochs} and
	\ref{fig:coco_comparison_heatmaps}, we give a comparison of performance for all
	considered approaches on the MS-COCO validation set. These learning curves and
	heatmaps allow us to appreciate the ability of RATT to remember old tasks. The
	forgetting rate of EWC seems to be higher than the one of LwF, but EWC shows an
	ability to recover performance after noticeable forgetting -- probably due to 
	increased backward transfer. This is clear looking at
	figure~\ref{fig:coco_comparison_heatmaps} in which both LwF and EWC seems to
	suffer noticeable forgetting on the first two tasks (transport and animal) after
	training on the third one (Sport). EWC seems able to recover when trained on the
	next task, while LwF continues to forget more.
	
	\begin{figure}[H]
		\includegraphics[width=1.0\linewidth]{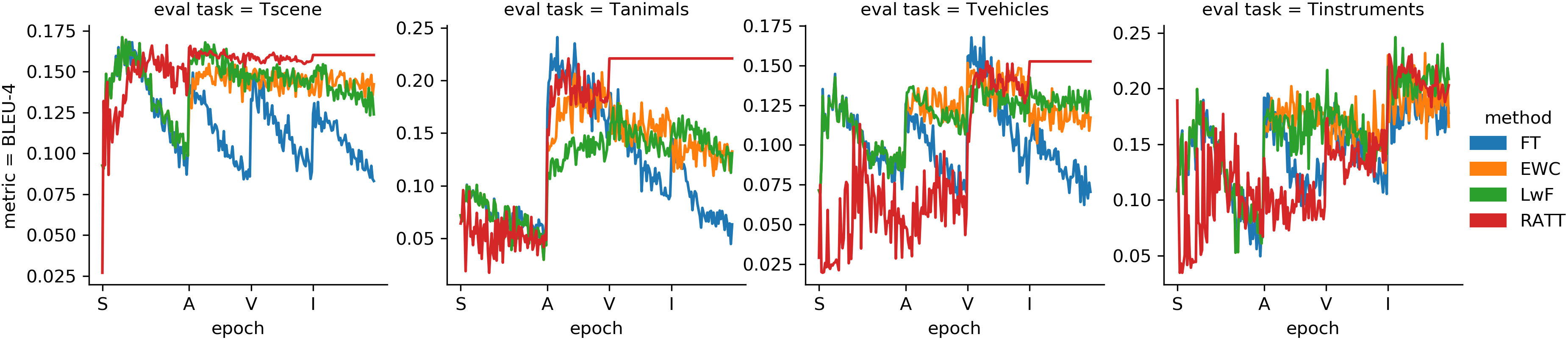}
		\vspace{-0.5cm}
		\caption{
			\label{fig:flickr_comparison_epochs}
			Comparison for all approaches on Flickr30k validations set. 
			Each plot reports BLEU-4 performance evaluated on one of the tasks at
			different training epochs and different training tasks for each of the
			ablated models. }
		\vspace{-0.3cm}
	\end{figure}
	\begin{figure}[H]
		\includegraphics[width=1.0\linewidth]{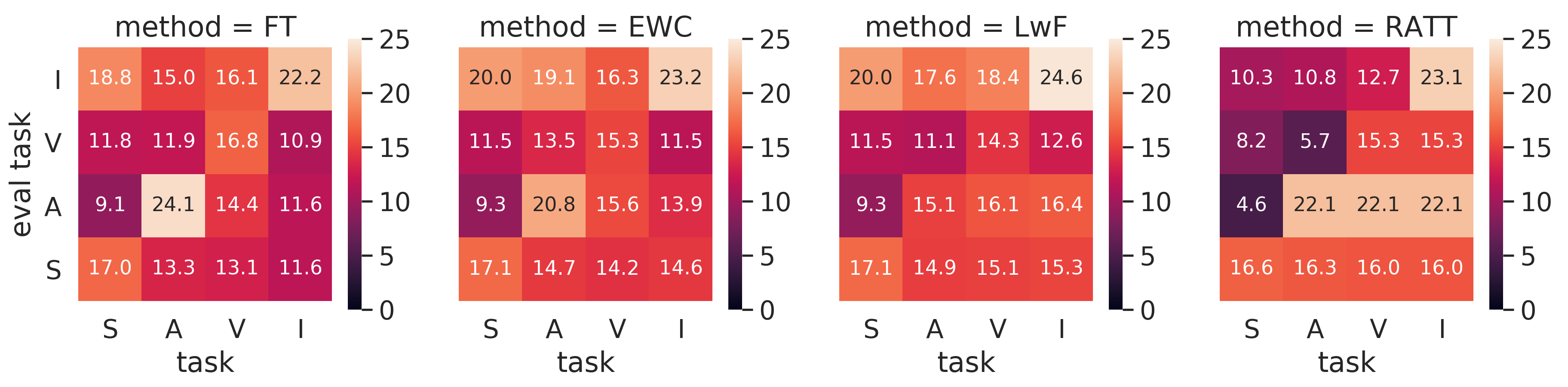}
		\vspace{-0.5cm}
		\caption{
			\label{fig:flickr_comparison_heatmaps}
			Comparison for all approaches on Flickr30k validations set. 
			Each heatmap reports BLEU-4 performance for one of the models evaluated on different tasks
			at the end of the training of each task.
		}
	\end{figure}
	
	\subsection{Learning and forgetting on Flickr30k}
	In figure \ref{fig:flickr_comparison_epochs} and
	\ref{fig:flickr_comparison_heatmaps} we give a comparison of performance for all
	approaches on the Flickr30k validation set. The first figure depicts the
	training process over all tasks, where the model is evaluated on each task while
	progressing through training. The results for Flickr30k show more variance than
	MS-COCO, as this setting is more challenging and the validation dataset is much
	smaller.
	
	RATT exhibits almost no forgetting in comparison to other methods -- an almost
	straight line after learning each task. Degradation of the FT model is visible,
	but for Flickr30k we notice that subsequent, more specific tasks keep previously
	learned and more generic concepts rather than completely forgetting (i.e. the
	first task category \textbf{scene}). The BLEU-4 score for LwF remains almost at
	the same level after learning the task, and EWC shows similar performance but
	with a bigger drop when going from task A to V. In
	figure~\ref{fig:flickr_comparison_heatmaps} an evaluation summary is provided in
	form of BLEU-4 heatmaps. Going from the left (FT) to right (RATT), less
	forgetting can be observed by each of evaluated method, with RATT showing almost
	no loss in performance when reaching the final task.
	
	It is useful to compare and contrast results on Flickr30k and MS-COCO. In
	Flickr30k there is much more information shared between tasks and this is shown
	by the significant forward transfer that we see: after training on the first
	task (scene), the performance on the last task (instrument) is significant for
	all methods. Forward transfer is much less evident for RATT, and this is due to
	the fact that it use the task embedding of future tasks for which it has no
	information (they all are randomly initialized). The backward transfer on
	Flickr30k is also evident looking at the relatively high performance of the FT
	baseline in figures \ref{fig:flickr_comparison_epochs} and
	\ref{fig:flickr_comparison_heatmaps} (and comparing with the MS-COCO equivalents
	in figures \ref{fig:coco_comparison_epochs} and
	\ref{fig:coco_comparison_heatmaps}).
	
	Although the overall performance on Flickr30k is much lower than on MS-COCO
	(evident when looking at the anti-diagonal of FT in figures
	\ref{fig:coco_comparison_heatmaps} and \ref{fig:flickr_comparison_heatmaps}),
	given the difficulty of the dataset itself and given the small number of
	examples (especially in validation/test sets) is difficult to draw firm
	conclusions about backward transfer for LwF and EWC.

	\section{Additional captioning results}
	\label{sec:results}
\begin{figure}[H]
		\includegraphics[width=1.0\linewidth]{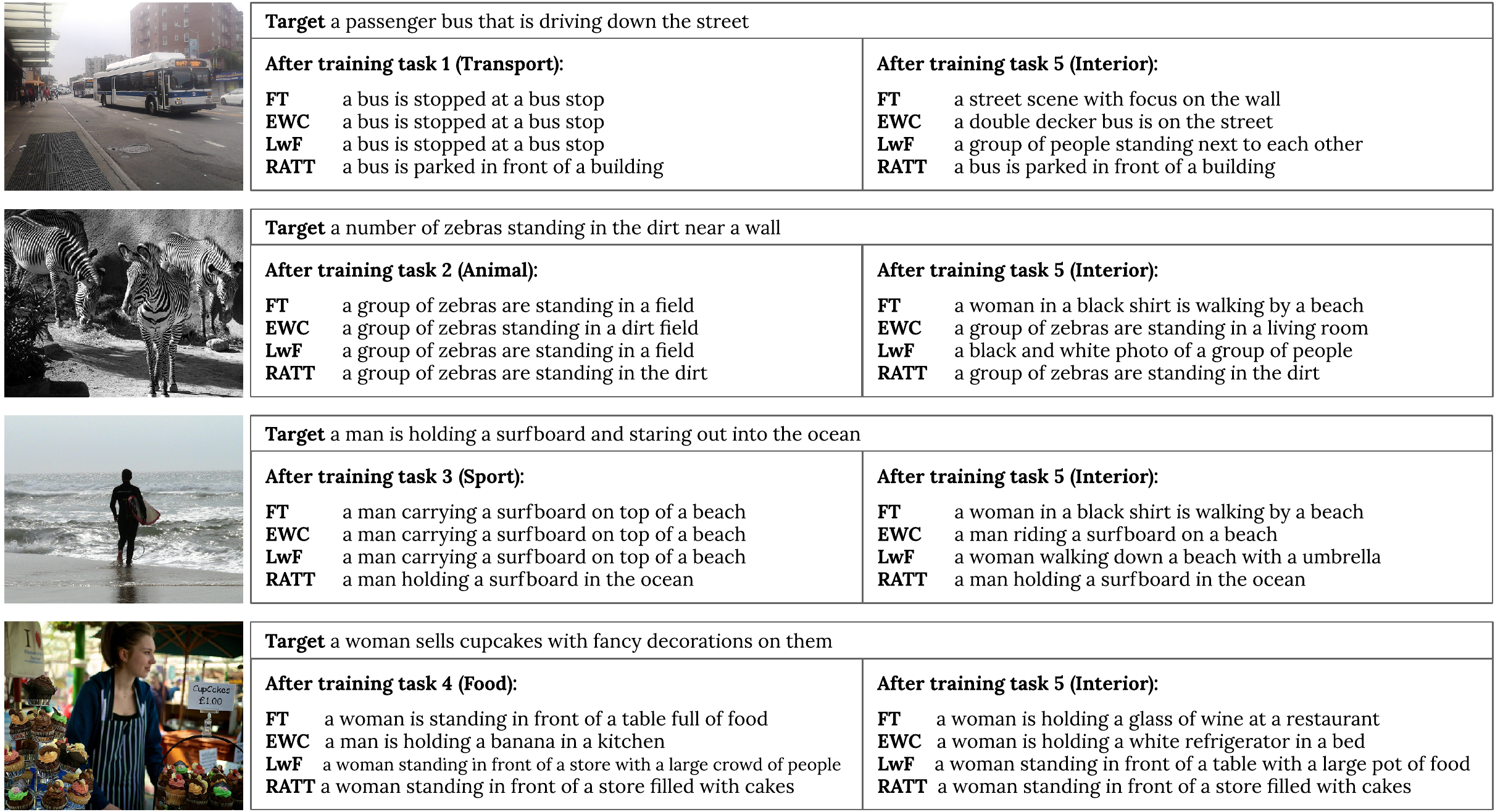}
		\vspace{-0.5cm}
		\caption{
			\label{fig:coco_qualitative}
			Captioning results for all methods on MS-COCO. Images and target captions
			belong to a specific task and captions are generated by all techniques after
			training the correct task (left) and a later task (right). Approaches except
			RATT contextualize to some degree generated captions with respect to the
			most recently learned task.}
	\end{figure}

\begin{figure}
		\includegraphics[width=1.0\linewidth]{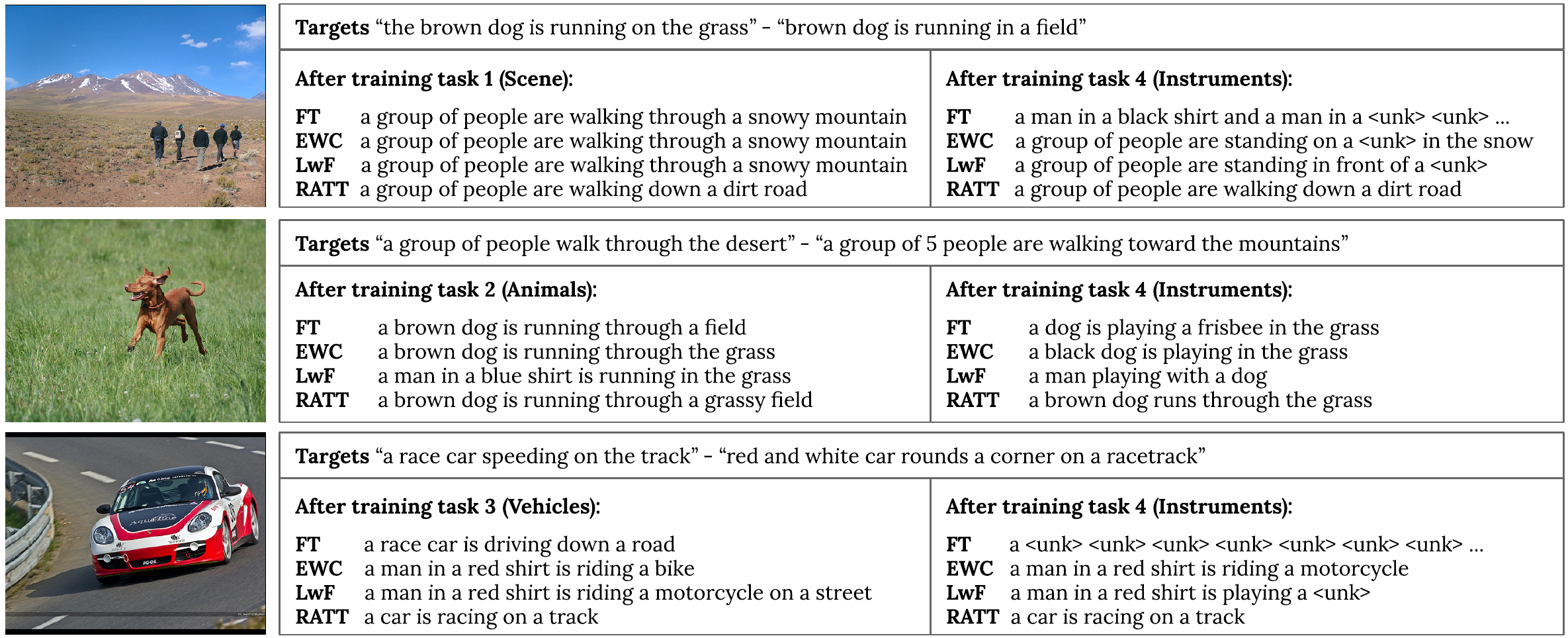}
		\vspace{-0.5cm}
		\caption{
			\label{fig:flickr_qualitative}
			Captioning results Flickr30k. Images and target captions belong to a 
			specific task and captions are generated by all techniques after
			training the correct task (left) and a later task (right).}
	\end{figure}

\begin{figure}
		\includegraphics[width=1.0\linewidth]{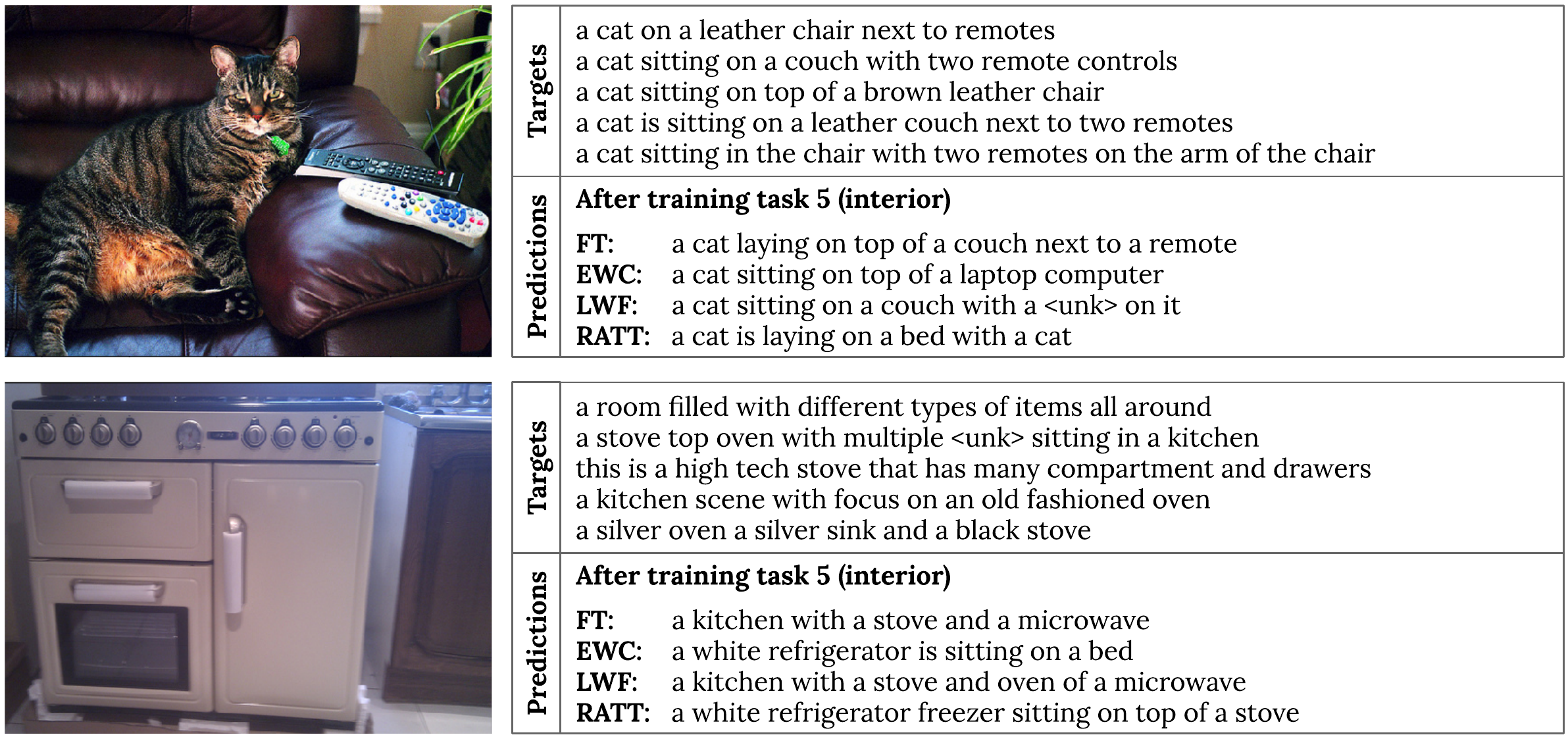}
		\vspace{-0.5cm}
		\caption{
			\label{fig:coco_failure}
			Examples of images from MS-COCO dataset for which fine tuning achieve better results than the proposed method. These images are taken from the last task, so there is no catastrophic interference. }
	\end{figure}

	In figure \ref{fig:coco_qualitative} we give an example image from each of the
	first four MS-COCO tasks with the prediction made by the models after
	training on the correct task (on the left) 
	and the one made
	after training on the complete sequence of tasks (on the right). Both EwC and LwF retain
	some correct words and ``insight'', but they are clearly confused by the last
	task on which they are trained. In the second image EWC predicts zebras in a
	living room because the last task contain house interiors. In a similar way, in
	the last picture EWC predicts the words \emph{refirgerator} and \emph{bed}, while
	LwF predicts \emph{table}. In figure \ref{fig:flickr_qualitative} we can see a
	similar analysis conducted on Flickr30k dataset. Again the quality of RATT
	captions is retained after training on the last task. 
	In figure
	\ref{fig:coco_failure} we give two qualitative examples taken from the last
	task from the MS-COCO dataset for which fine-tuning provides better descriptions than
	RATT. In this case the baseline does not suffer from catastrophic forgetting because 
	we evaluate the last trained task.
	RATT could be limited by the fact that neurons allocated to previous tasks are not trainable.
	
	
	% CHECK UNUSED REFERENCES with package \usepackage{refcheck}
%	\nocite{*}
	\clearpage

	\bibliography{supplementary}
	\bibliographystyle{plain}